\documentclass[journal]{IEEEtran}
\usepackage{times}
\usepackage{soul}
\usepackage{url}
\usepackage[hidelinks]{hyperref}
\usepackage[utf8]{inputenc}
\usepackage[small]{caption}
\usepackage{graphicx}
\usepackage{amsmath}
\usepackage{booktabs}
\usepackage{algorithm}
\usepackage{algorithmic}
\usepackage{lscape}

\usepackage{multirow}
\usepackage{subfigure}
\usepackage{amsfonts}
\usepackage{makecell}
\usepackage{xcolor}
\usepackage{multicol}
\usepackage{pdfpages}
\usepackage{standalone}

\ifCLASSINFOpdf
\else
\fi
\begin{document}
%
\title{Efficient Network Construction through Structural Plasticity}
%
%
%

\author{Xiaocong~Du,~\IEEEmembership{Student~Member,~IEEE,}
        Zheng~Li,~\IEEEmembership{Student~Member,~IEEE,}
        Yufei~Ma,~\IEEEmembership{Member,~IEEE,}
        and~Yu~Cao,~\IEEEmembership{Fellow,~IEEE}
\thanks{
Manuscript received April 21, 2019; revised July 1, 2019; accepted
July 31, 2019. Date of publication August 5, 2019; date of current version
September 17, 2019. This work was supported in part by the C-BRIC, one
of six centers in JUMP, in part by the Semiconductor Research Corporation (SRC) Program, and in part by the by the National Science Foundation (NSF) under CCF 1715443. This article was recommended by Guest Editor M. Ziegler. (Corresponding authors: Xiaocong Du; Yufei Ma.)

Xiaocong Du, Yufei Ma and Yu Cao are with the School of Electrical, Computer and Energy Engineering, Arizona State University, Tempe, AZ 85287, USA (e-mail: xiaocong@asu.edu; yufeima@asu.edu; ycao@asu.edu).  (Corresponding author: Xiaocong Du, Yufei Ma)}
\thanks{Zheng Li is with the School of Computing, Informatics, and Decision Systems Engineering, Arizona State University, Tempe, AZ 85287, USA  (e-mail: zhengl11@asu.edu). }
\thanks{Preprint version. For a full version, please refer to JETCAS.}
}
\maketitle

\begin{abstract}
Deep Neural Networks (DNNs) on hardware is facing excessive computation cost due to the massive number of parameters. A typical training pipeline to mitigate over-parameterization is to pre-define a DNN structure with redundant learning units (filters and neurons) with the goal of high accuracy, then to prune redundant learning units after training with the purpose of efficient inference. We argue that it is sub-optimal to introduce redundancy into training  in order to reduce redundancy later in inference. Moreover, the fixed network structure further results in poor adaption to dynamic tasks, such as lifelong learning. In contrast, structural plasticity plays an indispensable role in mammalian brains to achieve compact and accurate learning. Throughout the lifetime, active connections are continuously created while those that are no longer important are degenerated. Inspired by such observation, we propose a training scheme, namely Continuous Growth and Pruning (CGaP), where we start the training from a small network seed, then literally execute continuous growth by adding important learning units and finally prune secondary ones for efficient inference. The inference model generated from CGaP is sparse in the structure, largely decreasing the inference power and latency when deployed on hardware platforms. With popular DNN structures on representative datasets, the efficacy of CGaP is benchmarked by both algorithmic simulation and architectural modeling on Field-programmable Gate Arrays (FPGA). For example, CGaP decreases the FLOPs, model size, DRAM access energy and inference latency by 63.3\%, 64.0\%, 11.8\% and 40.2\%, respectively, for ResNet-110 on CIFAR-10.

\end{abstract}


%
\IEEEpeerreviewmaketitle


\section{Introduction}\label{sec:intro}

\begin{figure}[!t]
\begin{center}
\includegraphics[width=0.98\columnwidth]{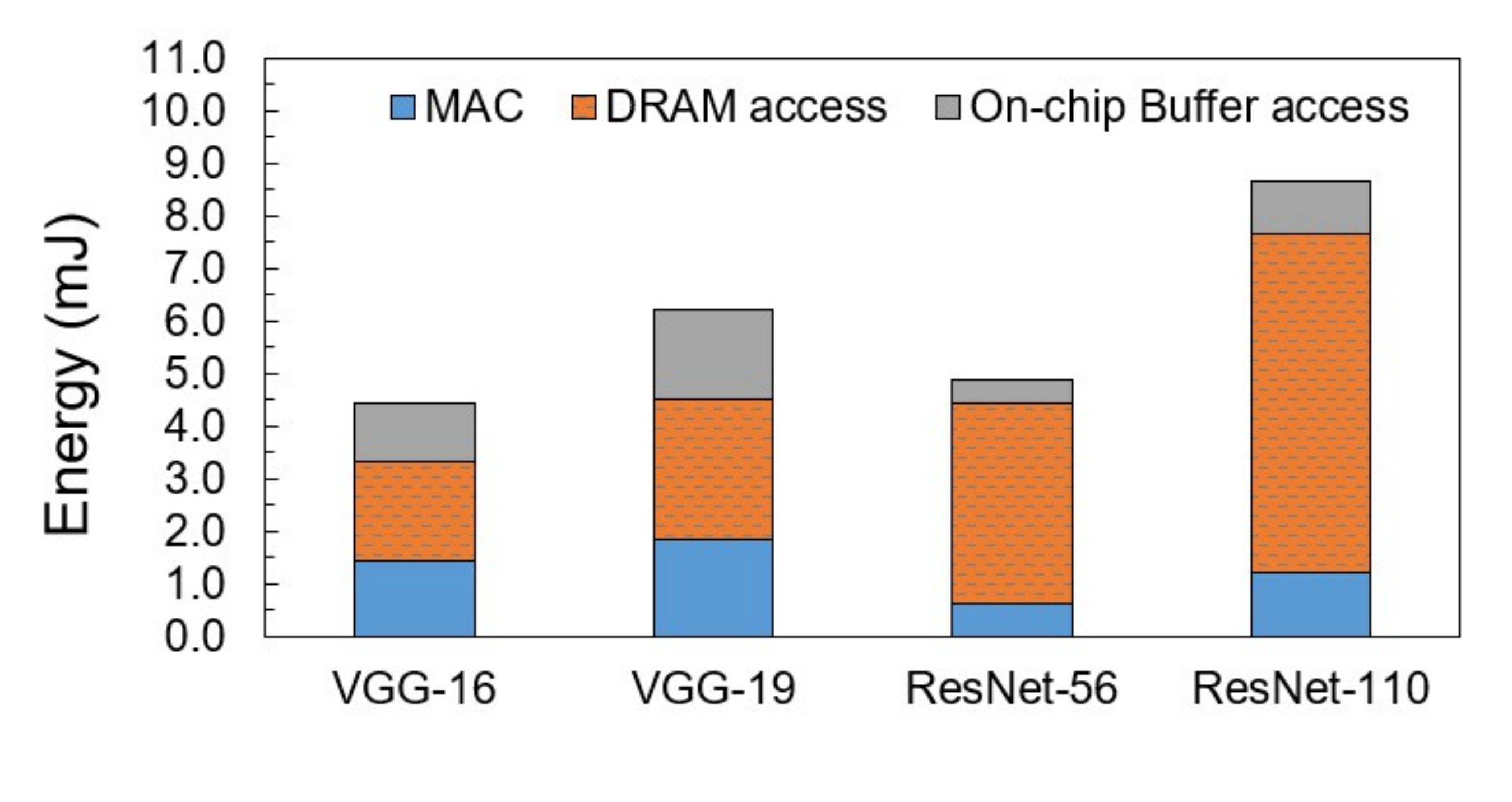}
\end{center}
\caption{Energy breakdown for modern DNN structures, results from simulation by the FPGA performance model~\cite{ma2019performance}. Due to the redundancy in parameters, multiply-accumulator (MAC) and external memory (DRAM) access dominate the energy consumption.}
\label{fig:45bar}
\end{figure}

\IEEEPARstart{D}{eep} Neural Networks have various applications including image classification \cite{krizhevsky2012imagenet}, object detection \cite{ren2015faster}, speech recognition \cite{graves2013speech} and natural language processing \cite{zhang2016yin}. 
However, the accuracy of DNNs  heavily relies on massive amounts of parameters and deep structures, making it hard to deploy DNNs on resource-limited embedded systems.  When training or inferring the DNN models on hardware, the model must be stored in the external memory such as dynamic random-access memory (DRAM) and fetched multiple times. These operations are expensive in computation, memory access, and energy consumption. For example, Fig.~\ref{fig:45bar} shows the energy consumption of one inference pass in several modern DNN structures, simulated by the FPGA performance model~\cite{ma2019performance} under the setting of 300 MHz operating frequency and 19.2 GB/s DRAM bandwidth. The input image size is $32\times32$. A typical DNN model is too large to fit in on-chip memory. For instance, VGG-19~\cite{simonyan2014very} has 20.4M parameters. Running such a model requires frequent external memory access, exacerbating the power consumption of a typical embedded system.

\begin{figure*}[!t]
\begin{center}
\includegraphics[width=0.98\textwidth]{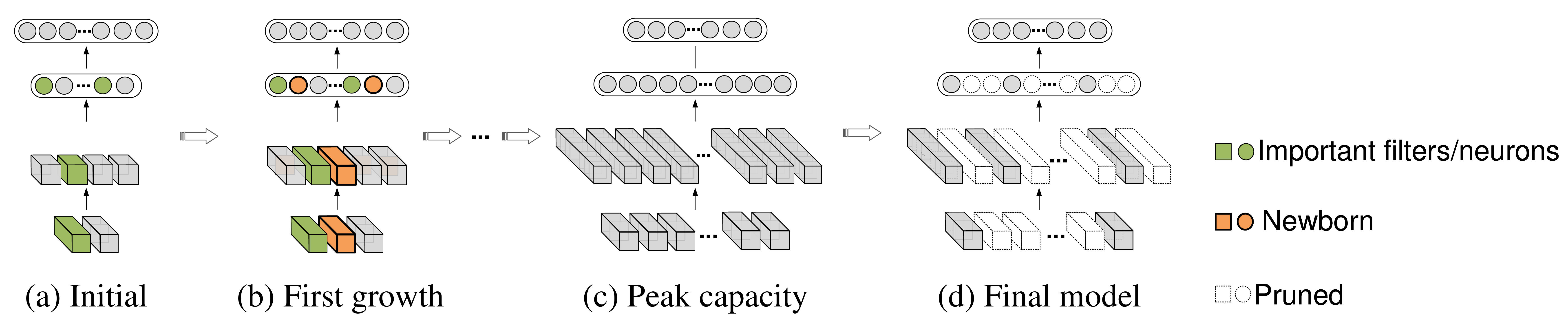}
\end{center}
\caption{The proposed CGaP scheme.  CGaP starts the training from a seed network instead of an over-parameterized one, gradually grows important learning units during the training and reaches peak capacity at the end of growth, then prunes secondary filters and neurons to generate an inference model with structured sparsity and up-to-date accuracy.}
\label{fig:flow}
\end{figure*}

Previous researches have designed customized hardware for DNN acceleration~\cite{chen2014dadiannao,du2015shidiannao}. 
Most of them are limited to relatively small neural networks, such as LeNet-5~\cite{lecun1998gradient}. For larger networks such as AlexNet~\cite{krizhevsky2012imagenet} and VGG-16~\cite{simonyan2014very}, additional efforts are usually required to improve the hardware efficiency~\cite{chen2016eyeriss,guo2017angel}. For example,  \cite{chen2016eyeriss} saves the energy through data gating and zero skipping.
Some other works focus on data reuse of convolutional layers and demonstrate the results on specific hardware~\cite{chen2014dadiannao,shafiee2016isaac,qiu2016going,farabet2009cnp}. However, their improvements are limited on those networks where fully-connected layer is widely used, such as RNNs and LSTMs.

To support more general models, network pruning is a popular approach by removing secondary weights and neurons. Network pruning executes a three-step procedure, which
1) trains a pre-designed network from scratch,
2) removes less important connections or filters/neurons according to a saliency score (a metrics to measure the importance of weights and learning units)~\cite{han2015learning,haoli,hu2016network,liu2017learning,luo2017thinet}, or by adding a regularization term into the loss function~\cite{lebedev2016fast,wen2016learning}, and 
3) fine-tunes to recover the accuracy.

However, the above pruning techniques suffer from two limitations: (1) training a large and fixed network from scratch could be sub-optimal as it introduces redundancy; (2) in the process of training, pruning only discards less important weights at the end of training but does not strengthen important weights and nodes. These limitations of network pruning confine the learning performance as well as the model pruning efficiency (i.e., how many parameters can be removed and how structured the sparsity is).

In contrast to the static DNN model, the biological nervous system exhibits active growth and pruning through the lifetime. \cite{gilmore2007regional, lipina2009poverty, butz2013simple} have observed that the rapid growth of neurons and synapses takes place in an infant's brain and is vital to the maturity of an adult's brain. In brains, some neurons and synapses are used more frequently and are consequently strengthened. Those neurons and synapses that are not used consistently are weakened and removed. The structural plasticity of brain is central to the study of developmental biology.

Inspired by this observation from biology, we propose a training scheme named Continuous Growth and Pruning (CGaP), which leverages structural plasticity to tackle the aforementioned limitations of pruning techniques. 
Instead of training an over-parameterized network from scratch, CGaP starts the training from a small network seed (Fig.~\ref{fig:flow}(a)), whose size is as low as 0.1\%-3\%  of the full-size reference model. In each iteration of the growth,  CGaP locally sorts neurons and filters (also known as output channels in some literature) according to our saliency score (Section~\ref{section:score}). Based on the saliency score, important learning units are selected and the corresponding new units are added (see Fig.~\ref{fig:flow}(b)). The selection and addition of important units help reinforce the learning and increase model capacity. Then a filter-wise and neuron-wise pruning will be executed on the post-growth model (Fig.~\ref{fig:flow}(c)) based on pruning metrics. Finally, CGaP generates a significantly sparse and structured inference model (Fig.~\ref{fig:flow}(d)) with accuracy improved.  
In the generated inference model, large amounts of filters and neurons have been removed, achieving structured pruning. Compared to non-structured pruning~\cite{han2015learning}, CGaP benefits hardware implementation as it reduces the computation volume and memory access without any additional hardware architecture change.

Algorithmic experiments and hardware simulations validate that CGaP significantly decreases the number of external and on-chip memory accesses, accelerating the inference by bypassing the removed filters and neurons. On the algorithm side, we demonstrate the performance in accuracy and model pruning on several networks and datasets. For instance, CGaP reduces $78.9\%$ parameters of VGG-19 with $+0.37\%$ accuracy improvement on CIFAR-100~\cite{krizhevsky2009learning}, $85.8\%$ parameters with $+0.23\%$ accuracy improvement on SVHN~\cite{netzer2011reading}. For ResNet-110~\cite{he2016deep}, CGaP reduces $64.0\%$ parameters with $+0.09\%$ accuracy improvement on CIFAR-10~\cite{krizhevsky2009learning}. These results exceed the state-of-the-art pruning methods~\cite{han2015learning,haoli,hu2016network,liu2017learning,liu2018rethinking,he2018soft}. Furthermore, we validate the efficiency of the inference model generated from CGaP using FPGA simulator~\cite{ma2019performance}. For one inference pass of VGG-19 on CIFAR-100, previous non-structured pruning approach~\cite{han2015learning} requires energy consumption of $2.7\times10^9$ pJ in accessing DRAM and 5.6 ms inference latency, while CGaP requires only $2.2\times10^9$ pJ and 4.4 ms latency.

The contribution of this paper is as follows:
\begin{itemize}
\item A brain-inspired training flow (CGaP) with a dynamic structure is proposed. CGaP grows the network from a small seed and effectively reduces over-parameterization without sacrificing  accuracy. 
\item The advantage of structured sparsity of the inference model generated from CGaP is validated using a high-level FPGA performance model including on-chip buffer access energy, external memory access energy and inference latency.
\item The discussion and understanding of the reason that the growth improves the learning efficiency are provided. 
\end{itemize}

The rest of the paper is organized as follows. Section~\ref{sec:related} introduces the background of model pruning. Section~\ref{sec:Saliency} demonstrates the saliency score used to select the learning units. Section~\ref{sec:method} describes the proposed Continuous Growth and Pruning scheme. Section~\ref{sec:SWresult} presents the experimental results from algorithmic simulations. Section~\ref{sec:HWresult} demonstrates the simulation results from FPGA performance modeling. Section~\ref{section:discussion} discusses the understanding of network plasticity as well as ablation study. Section~\ref{sec:conclusion} concludes this work and discusses the insight into future work.

\section{Previous Work}
\label{sec:related}

There have been broad interests in reducing the redundancy of DNNs in order to deploy them on a resource-limited hardware platform. The structural surgery is a widely used approach and can be categorized into destructive direction and constructive direction. We will discuss these two directions, as well as orthogonal approaches to our methods in this section.

\subsection{Destructive Methods}
The destructive methods zero out specific connections or remove filters or neurons in convolutional or fully-connected layers,  generating a sparse model. Weight magnitude pruning~\cite{han2015learning} pruned weights by setting the selected weights to zeros. The selection is based on L1-norm, i.e., the absolute value of the weight. Weight magnitude pruning generates a sparse weight matrix, but not in a structured way. In this case, specific hardware design~\cite{han2016eie} is needed to take advantage of the optimized inference model, otherwise the non-structured sparsity does not benefit hardware acceleration due to the overhead in model management. The kernel-wise pruning~\cite{haoli} pruned kernels layer by layer based on the saliency metrics of each filter and achieved structured sparsity in the inference model. Compared to~\cite{haoli}, CGaP prunes filters, leading to a more structured inference model. Besides the saliency-based pruning, the penalty-based approach has been explored by~\cite{wen2016learning,liu2015sparse} and structured sparsity was achieved. Our method is different from all the above pruning schemes from two perspectives: (1) We start training from a small seed other than an over-parameterized network; (2) Besides removing secondary filters/neurons, we also reinforce important ones to further improve learning accuracy and model compactness.

\subsection{Constructive Methods}
The constructive approaches include techniques that add new connections or filters to enlarge the model capacity. \cite{ash1989dynamic,briedis1998using} increased network size by adding random neurons with fresh initialization (i.e., weights are randomly initialized, without pre-trained information). They evaluated their approach on basic XOR problems. Different from their approach, CGaP selectively adds neurons and filters that are initialized with the information learned from the previous training. Meanwhile, CGaP is validated on modern DNNs and datasets under more realistic scenarios. 
\cite{sakar1993growing} grew the smallest Neural Tree Networks (NTN) to minimize the number of classification errors on Boolean function learning tasks, and used pruning to enhance the generalization of NTN.
~\cite{huang2005generalized} improved the accuracy of radial basis function (RBF) networks on function approximation tasks by adding and removing hidden neurons.
To enhance the accuracy of spike-based classifiers, ~\cite{hussain2016multiclass} progressively added dendrites to the network, and then optimized the topology of the dendritic tree.  Different from them, CGaP aims at improving the efficiency of the inference model of modern Deep Neural Networks on image classification tasks.
~\cite{dai2017nest} constructed the DNN by activating connections and choosing a set of convolutional filters among a bunch of randomly generated filters according to their influence on the training performance. However, this approach highly depends on trial and error to find the optimal set of filters that could reduce loss the most. This approach is sensitive to power and timing budgets, limiting its extension on large datasets. Unlike their work, CGaP directly grows the network from a seed, minimizing the effort on trail and error.

\subsection{Orthogonal Methods}

The orthogonal methods, such as low-precision quantization and low-rank decomposition,  compress the DNN models by quantizing the parameters to fewer bits~\cite{gong2014compressing,hubara2017quantized}, or by finding a low-rank approximation~\cite{denton2014exploiting,leng2018extremely}. Note that our CGaP approach can be combined with these orthogonal methods to further improve inference efficiency.


\section{Saliency Score}
\label{sec:Saliency}
In this section, we describe the detailed methodology of CGaP, starting from the saliency score, which is used to sample the importance of a learning unit. Section~\ref{section:terminology} defines the terminology we use in this paper. Section~\ref{section:score} provides the mathematical proof of the saliency score we adopt.

\subsection{Terminology}
\label{section:terminology}
 
A DNN can be treated as a feedforward multi-layer architecture that maps the input images to certain output vectors. Each layer is a certain function, such as convolution, ReLU, pooling and inner product, whose input is $\mathcal{X}$, output is $\mathcal{Y}$ and parameter is $\mathcal{W}$ in case of convolutional and fully-connected layers. Hereby the convolutional layer (conv-layer) is formulated as: $\mathcal{Y}_l = \mathcal{X}_l\ast\mathcal{W}_l$, wherein $ \mathcal{X}_l \in \mathbb{R}^{I_l \times Wi_l \times Hi_l}$, $\mathcal{Y}_l \in \mathbb{R}^{O_l \times Wo_l \times Ho_l}  \Leftrightarrow \mathcal{X}_{l+1} \in \mathbb{R}^{I_{l+1} \times Wi_{l+1} \times Hi_{l+1}}$, $
 \mathcal{W}_l \in \mathbb{R}^{O_l \times I_l \times K \times K}$, where subscript $_{l}$ denotes the index of the layer.  And the fully-connected layer is represented by: $\mathcal{Y}_l = \mathcal{X}_l \cdot \mathcal{W}_l$,    where the input $\mathcal{X}_l \in \mathbb{R}^{I_l}$, the output $\mathcal{Y}_l \in \mathbb{R}^{O_l}     \Leftrightarrow \mathcal{X}_{l+1} \in \mathbb{R}^{I_{l+1}}$, and the parameter matrix is $\mathcal{W}_l \in 
    \mathbb{R}^{O_l \times I_l}$.

\paragraph{Convolutional layer (conv-layer) $l$} the 4 dimensions of its weight matrix are: the number of output channels $O_l$, the number of input channels $I_l$, and the kernel width and height $K$, respectively. We denote the $o$-th \textbf{3D filter}, which generates the $o$-th output channel in the feature map, as $W^o_l\in\mathbb{R}^{I_l\times K \times K}$. The $i$-th \textbf{2D kernel} in the $o$-th filter is denoted as $W_l^{o,i}\in\mathbb{R}^{K\times K}$. 
On the other hand, a \textbf{4D weight tensor} $\mathbf{W}_l^i\in\mathbb{R}^{O_l\times 1\times K\times K}$ , which operates on the $i$-th input feature map, is a package of $O_l$ kernels across all output channels. For example, in Fig.~\ref{fig:layer_wise_a}, $W_{l, picked}^j$ is a 3D filter consisting of $I_l$ kernels, and $\mathbf{W}_{l+1, projected}^j$ as well as $\mathbf{W}_{l+1, mapped}^j$ are both 4D tensors with dimension of $O_l\times 1\times K\times K$, which include all the output channels but have only one input channel located at j. 
The $W_l^{o,i,m,n}\in\mathbb{R}^{1\times1}$ refers to one weight at the $m$-th row and the $n$-th column in the $o$-th filter of the $i$-th input channel. 

\paragraph{Fully-connected layer (fc-layer) $l$} input $\mathcal{X}_l$ propagate from one hidden activation $i$ to the next layer. We refer the whole set of $W^i_{l,fan-out}$ as a neuron $N^i_l$. This neuron receives information from previous layer $l-1$ through its \textbf{fan-in} weights $W_{l,fan-in}^i\in\mathbb{R}^{1\times I_{l-1}}$ (as shown in Fig.~\ref{fig:layer_wise_b}) and propagates to the next layer through \textbf{fan-out} weights $W^i_{l,fan-out}\in\mathbb{R}^{O_l\times 1}$. Also note that  the output dimension of layer $l-1$ equals to the input dimension of layer $l$, i.e., $O_{l-1}=I_{l}$. The weight pixel in layer $l$ at the cross-point of row $o$ and column $i$ is denoted as $W^{o,i}_{l,fan-out}\in\mathbb{R}^{1\times 1}$.
Moreover,  the `depth' of a DNN model indicates the number of layers, and the `width' of a DNN model refers to the number of filters or neurons of each layer.

\paragraph{Learning units} Growing or pruning a \textbf{filter} $W^o_l$ indicates adding or removing $W^o_l\in\mathbb{R}^{I_l\times K\times K}$ and its corresponding output feature map. Growing or pruning a \textbf{neuron}  $N^i_l$ means adding or removing both $W^i_{l,fan-out}\in\mathbb{R}^{O_l\times 1}$ and $W_{l,fan-in}^i\in\mathbb{R}^{1\times I_{l-1}}$.

\begin{algorithm}[!t]
\caption{Entire flow}
\label{CGaP-alg1}
\textbf{Input}: Model seed $M_{initial}$
\begin{algorithmic}[1] 
\STATE Initialize a small network model $M_{current}\leftarrow M_{initial}$.
\FOR{epoch = 1 to E}
    \STATE Train current model $M_{current}$ and fetch $Accuracy$.
    \IF {$epoch\%\frac{1}{f_{growth}}=0$ and $M_{current}$ \textless $\tau_{capa.}$}
      \STATE Grow the network according to Algorithm~\ref{CGaP-alg2}
      \STATE $M_{current} \leftarrow M_{grown}$. 
    \ENDIF
    \STATE $M_{peak} \leftarrow M_{current}$.
    \IF {epoch$ \% \frac{1}{f_{pruning}}=0$ and $Accuracy>\tau_{accu.}$}
        \STATE Prune the network following Algorithm~\ref{CGaP-alg3}
        \STATE $M_{current} \leftarrow M_{pruned}$.
    \ENDIF
\ENDFOR
\STATE $M_{final} \leftarrow M_{current}$ and test $M_{final}$.
\end{algorithmic}
\textbf{Output}: Final compact model $M_{final}$ 
\end{algorithm}

\begin{algorithm*}[!t]
\caption{Growth phase}
\label{CGaP-alg2}
\textbf{Input}: Current network $M_{current}$
\begin{algorithmic}[1] 
      \FOR{each layer $l$ = 1 to L}
            \FOR{each filter $W^o_l$ in conv-layer $l$, or each neuron $N^i_l$ in fc-layer $l$}
             \STATE  Calculate growth score $GS_{W^o_l}$ according to Eq.~\ref{math:filter_score} and $GS_{N^i_l}$ according to Eq.~\ref{math:neuron_score}.
            \ENDFOR
            \STATE Sort all units and select $\beta O_l$ filters or $\beta I_l$ neurons with the highest $GS_{W^o_l}$ or $GS_{{N^i_l}}$.
            \FOR{each filter j = 1 to $\beta O_l$ (for fc-layer, $\beta I_l$)}
                \STATE Add one filter/neuron on the side of the each picked filter/neuron in layer $l$.
                \STATE Initialize picked and new-born filters (neurons) according to Eq.~\ref{math:initial1} and Eq.~\ref{math:initial2}.
                \STATE Map corresponding input-wise weight in layer $l+1$ (fan-in weights in layer $l-1$).
                \STATE Initialize projected and mapped filters according to Eq.~\ref{math:initial3} and Eq.~\ref{math:initial4} (neurons according to Eq.~\ref{math:initial3-fc} and Eq.~\ref{math:initial4-fc}).
            \ENDFOR
      \ENDFOR
\end{algorithmic}
\textbf{Output}: $M_{grown}$ 
\end{algorithm*}

\begin{algorithm*}[!t]
\caption{Pruning phase}
\label{CGaP-alg3}
\textbf{Input}: Current network $M_{current}$
\begin{algorithmic}[1] 
\FOR{each weight $W_l^{o, i, m, n}\in\mathbb{R}^{1\times1}$ in conv-layer $l$ or each $W^{o,i}_l\in\mathbb{R}^{1\times1}$ in fc-layer $l$}
        \STATE Calculate weight pruning score $PS_W$ according to Eq.~\ref{math:PS1} for conv-layers and Eq.~\ref{math:PS2} for fc-layers.
        \ENDFOR
        \STATE Sort weights by $PS_W$.
        \STATE Zero-out the lowest $\gamma_W \prod(O_l, I_l, K, K)$ weights in conv-layer and $\gamma_W \prod(I_l, O_l)$ weights in fc-layer.
        \FOR{each filter $W^o_l$ (neuron $N^i_l$) in all layers}
           \STATE Zero-out entire filter $W^o_l$ (neuron $N^i_l$) if the weight sparsity is larger than pruning rate $\gamma_F$ ($\gamma_N$).
\ENDFOR
\end{algorithmic}
\textbf{Output}: $M_{pruned}$ 
\end{algorithm*}

\subsection{Saliency Score} 
\label{section:score}

We adopt a saliency score to measure the effect of a single filter/neuron on the loss function, i.e., the importance of each learning unit. The saliency score is developed from Taylor Expansion of the loss function. Previously, ~\cite{molchanov2016pruning} applied it on pruning. In this paper, we adopt this saliency score and apply it on the growth and pruning scheme. In this section, we provide a mathematical formulation of the saliency score.

The saliency score represents the difference between the loss with and without each unit. In other words, if the removal of a filter/neuron leads to relatively small accuracy degradation, this unit is recognized as an unimportant unit, and vice versa. Thus, the objective function to get the filter with the highest saliency score is formulated as:
\begin{align}
   \underset{W_l^o} {argmin} & {|\Delta\mathcal{L}(W_l^o)|}  \Leftrightarrow  \nonumber \\
  & \underset{W_l^o} {argmin} |\mathcal{L}(\mathcal{Y};\mathcal{X},\mathcal{W})  -\mathcal{L}(\mathcal{Y};\mathcal{X},W_l^o=\mathbf{0})|.
\end{align} 

Using the first-order of the Taylor Expansion:
\begin{align}
|\mathcal{L}(\mathcal{Y};\mathcal{X},\mathcal{W})-\mathcal{L}(\mathcal{Y};\mathcal{X},W_l^o=\mathbf{0})| \  at\  W_l^o = \mathbf{0}.
\end{align} 
we get:
\begin{align}\label{math:filter_score}   
   |\Delta\mathcal{L}(W_l^o)| &\simeq |\frac{\partial{\mathcal{L}}(\mathcal{Y};\mathcal{X},\mathcal{W})}{\partial{W_l^o}}W_l^o|  \nonumber\\ 
   & =\sum_{i=0}^{I_l}\sum_{m=0}^{K}\sum_{n=0}^{K}|\frac{\partial{\mathcal{L}}(\mathcal{Y};\mathcal{X},\mathcal{W})}{\partial{W_l^{o, i, m, n}}} W_l^{o, i, m, n}|. 
\end{align} 

Similarly, the saliency score of a neuron is derived as:
\begin{eqnarray} \label{math:neuron_score}  
   |\Delta\mathcal{L}(N_l^i)| &\simeq& |\frac{\partial{\mathcal{L}}(\mathcal{Y};\mathcal{X},\mathcal{W})}{\partial{W_{l,fan-out}^i}}W_{l,fan-out}^i|  \nonumber\\  \textbf{}
   &=& \sum_{o=0}^{O_l}|\frac{\partial{\mathcal{L}}(\mathcal{Y};\mathcal{X},\mathcal{W})} {\partial{W_{l,fan-out}}^{o, i}} W_{l,fan-out}^{o, i}|. 
\end{eqnarray}


\section{CGaP Methodology}
\label{sec:method}

With the saliency score as the foundation, we develop the entire CGaP flow atop. This section explains the overall flow and the detailed implementation of each step in CGaP. 

The CGaP scheme is described in Algorithm~\ref{CGaP-alg1}. Starting from a small network seed, the growth takes place periodically at a frequency of $f_{growth}$ 
(see Algorithm~\ref{CGaP-alg1} line 4, where `\%' denotes the operation to obtain the remainder of division). During each growth, important learning units are chosen and grown at growth ratio $\beta$ layer by layer from the bottom (input) to top (output), based on the local ranking of the saliency score. The growth phase stops when reaching a capacity threshold $\tau_{capa.}$, followed by several epochs of training on the peak model $M_{peak}$. When the training accuracy reaches a threshold $\tau_{accu.}$, the pruning phase starts. Pruning is performed layer by layer, from the bottom layer to the top layer, at the frequency of $f_{pruning}$. The details in the growth phase and the pruning phase is demonstrated as follows.

\subsection{Growth phase}\label{sec:growth}
\begin{figure}[t!]
\begin{center}
\includegraphics[width=\columnwidth]{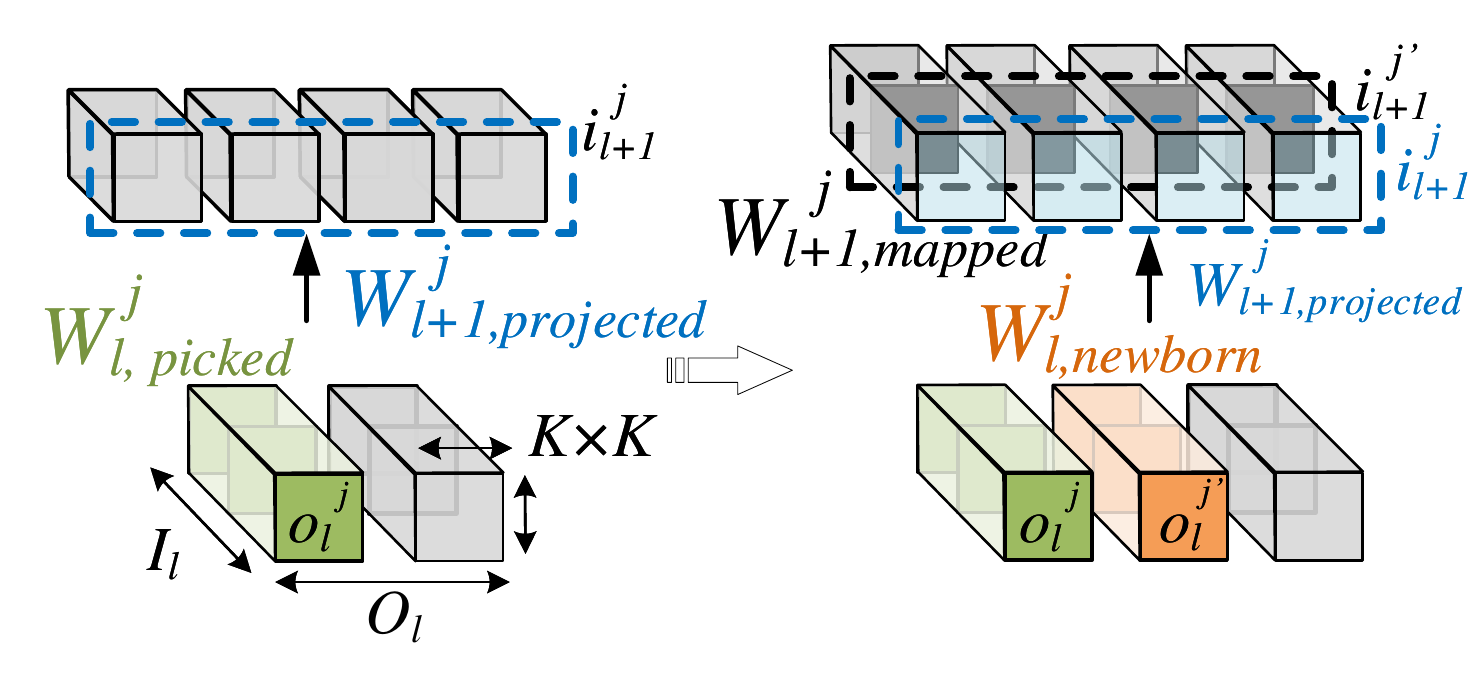}
\end{center}
\caption{Illustration of two-step growth in conv-layers. The growth phase follows a two-step (growing and mapping) procedure. After the filter $W_{l, picked}^j$ (green) is picked and split aside, giving birth to $W_{l, newborn}^j$ (orange), the projected input-wise filter,  $\mathbf{W}_{l+1, projected}^j$ (blue) in layer $l+1$, is as well split aside, generating $\mathbf{W}_{l+1, mapped}^j$ (black). }
\label{fig:layer_wise_a}
\end{figure}

\begin{figure}[t!]
\begin{center}
\includegraphics[width=0.65\columnwidth]{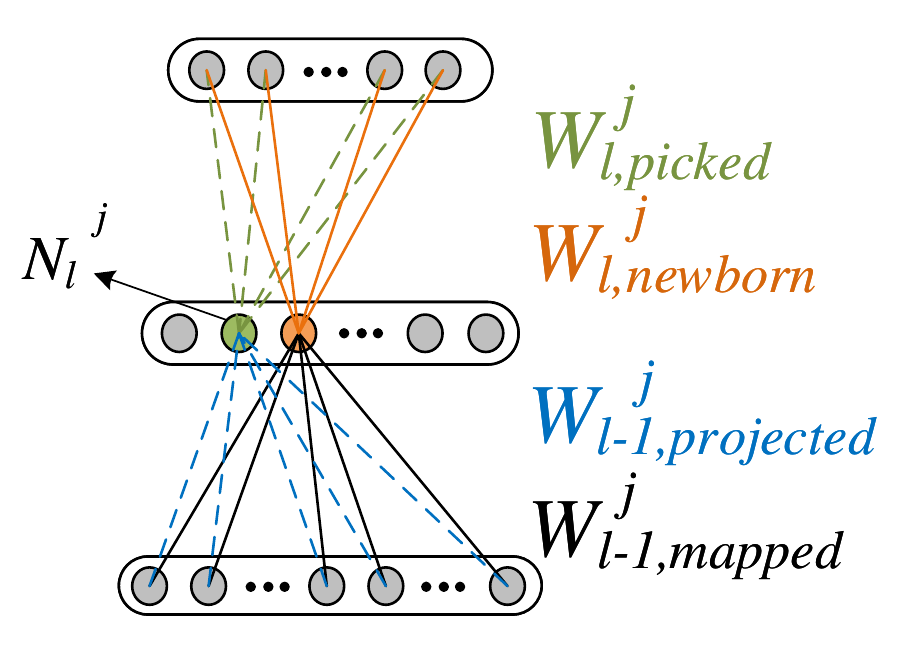}
\end{center}
\caption{Illustration of two-step growth in fc-layers. First, fan-out weights $W_{l, newborn}^j$ (orange) is added, then fan-in weights $\mathbf{W}_{l-1,mapped}^j$ (black) form the connections from the newborn neuron to all neurons in layer $l-1$.}
\label{fig:layer_wise_b}
\end{figure}

Algorithm~\ref{CGaP-alg2} presents the methodology in the growth phase. Each iteration of growth in a layer consists of two steps: growth in layer $l$ and mapping in the adjacent layer. There are two conditions need to be discussed separately: convolutional layers (Fig.~\ref{fig:layer_wise_a}) and fully-connected layers (Fig.~\ref{fig:layer_wise_b}). Due to the difference between these two kinds of operation as discussed previously, after the growth of layer $l$, the mapping in conv-layer takes place at the adjacent layer $l+1$. In fc-layers, the mapping is in layer $l-1$.

\paragraph{Growth in conv-layer $l$} 
According to the local ranking of the saliency score (Eq.~\ref{math:filter_score}), we sort all the 3D filters in this layer. With a growth ratio $\beta$,  $\beta O_{l,t}$ filters are selected in the $l$-th layer at the $t$-th growth. On the side of each selected filter $\mathbf{W}_{l, picked}^j\in\mathbb{R}^{I_l\times K\times K}$, as shown in Fig.~\ref{fig:layer_wise_a}, we create a new filter that has the same size, named $\mathbf{W}_{l, newborn}^j\in\mathbb{R}^{I_l\times K\times K}$.

In the ideal case, the new filter $\mathbf{W}_{l, newborn}^j$ and existing filter $\mathbf{W}_{l, picked}^j$ are expected to collaborate with each other and optimize the learning. The existing filter $\mathbf{W}_{l, picked}^j$ has already learned on the current task. To keep the same learning pace between the existing filter and the new filter, we initialize $\mathbf{W}_{l, newborn}^j$ as follows: 
\begin{align}
\label{math:initial1}
\mathbf{W}_{l, newborn}^j = \sigma \mathbf{W}_{l, picked}^j + X \sim U([-\mu, \mu]), \\
\label{math:initial2}
\mathbf{W}_{l, picked}^j = \sigma \mathbf{W}_{l, picked}^j + X \sim U([-\mu, \mu]),
\end{align}
where $\sigma$ $\in(0,1]$ is a scaling factor and $X$ is a constant following uniform distribution in $[-\mu, \mu]$, where $\mu$ $\in(0,1]$. 
Instead of random initialization, the above initialization helps reconcile the learning status of the newborn filters with the old filters. Meanwhile, the scaling factor prevents output from an exponential explosion caused by the feedforward propagation $\mathcal{Y}_l = \mathcal{X}_l\ast\mathcal{W}_l$. The noise $X$ prevents the learning from sticking at a local minimum that leads to sub-optimal solutions. 
No matter which distribution the noise $X$ follows, $X$ in a reasonable range is able to provide similar performance. However, other distributions usually introduce more hyper-parameters and thus, require more efforts in parameter tuning. For example, Gaussian noise introduces more hyper-parameter, e.g., the standard deviation, than uniform noise. For simplicity, we use uniformly distributed noise.

\paragraph{Mapping in conv-layer $l+1$}
After the number of filters in layer $l$ grows from $O_{l,t}$ to $(1+\beta)O_{l,t}$, the number of output feature maps also increases from $O_{l,t}$ to $(1+\beta)O_{l,t}$. Therefore, the input-wise dimension of layer $l+1$ should increase correspondingly in order to be consistent in data propagation.
To match the dimension, we first locate the 4D tensor $\mathbf{W}_{l+1, projected}^j$ in layer $l+1$, which processes the feature maps generated by $W_{l, picked}^j$. Then we add a new 4D tensor $\mathbf{W}_{l+1, mapped}^j$ adjacent to $\mathbf{W}_{l+1, projected}^j$.
The $\mathbf{W}_{l+1, mapped}^j$ and $\mathbf{W}_{l+1, projected}^j$ are initialized as follows:
\begin{align}
\label{math:initial3}
\mathbf{W}_{l+1, mapped}^j = \sigma \mathbf{W}_{l+1, projected}^j + X \sim U([-\mu, \mu]), \\
\label{math:initial4}
\mathbf{W}_{l+1, projected}^j = \sigma \mathbf{W}_{l+1, projected}^j + X \sim U([-\mu, \mu]).
\end{align}

To summarize, as illustrated in Fig.~\ref{fig:layer_wise_a}, the filter $W_{l, picked}^j$ (green) is selected according to the saliency score and  a new tensor $W_{l, newborn}^j$ (orange) is added. Then the input-wise tensor $\mathbf{W}_{l+1, projected}^j$ (in blue dashed rectangular) in layer $l+1$ is projected, and $\mathbf{W}_{l+1, mapped}^j$ (in black dashed rectangular) is generated.
 
After layer $l$ grows and layer $l+1$ is mapped, layer $l+1$ grows and layer $l+2$ is mapped, so on and so forth till the last convolutional layer. It is worth mentioning that for the `projection shortcuts'~\cite{he2016deep} with $1\times1$ convolutions in ResNet~\cite{he2016deep}, the dimension mapping is between the two layers that the shortcut connects, not necessarily to be the adjacent layers. 

\paragraph{Growth and mapping in fc-layers}
As illustrated in Fig.~\ref{fig:layer_wise_b}, the neuron growth in fc-layers $l$ occurs at fan-out weights, and its initialization follows Eq.~\ref{math:initial1} and \ref{math:initial2}. 

The mapping in fc-layers take place in the fan-in weights as follows: 
\begin{align}
\label{math:initial3-fc}
\mathbf{W}_{l-1,mapped}^j= \sigma \mathbf{W}_{l-1,projected}^j + X \sim U([-\mu, \mu]), \\ 
\label{math:initial4-fc}
\mathbf{W}_{l-1,projected}^j= \sigma  \mathbf{W}_{l-1,projected}^j + X \sim U([-\mu, \mu]).
\end{align}

After growing the last conv-layer, We flatten the output feature map of this conv-layer, treat it as the input from layer $l-1$ and map in the same manner.

\subsection{Pruning phase}
Pruning in each layer consists of two steps: weight pruning and unit pruning. First, we sort weight pixels locally in each conv-layer according to Eq.\ref{math:PS1}:
\begin{align}
\label{math:PS1}
PS_{W_l^{o, i, m, n}} = & |\frac{\partial{\mathcal{L}}(\mathcal{Y};\mathcal{X},\mathcal{W})}{\partial{W^{o, i, m, n}_l}}W_l^{o, i, m, n}|  
\end{align}
and in each fc-layer according to Eq.\ref{math:PS2}: 
\begin{align}
\label{math:PS2}
PS_{W_l^{o,i}} = &|\frac{\partial{\mathcal{L}}(\mathcal{Y};\mathcal{X},\mathcal{W})}{\partial{W_{l,fan-out}}^{o,i}}W_{l,fan-out}^{o,i}|
\end{align}
In each layer, 100$\gamma_W$\% weight pixels with the lowest $PS_W$ are set as zero, where $\gamma_W \in (0,1)$ is the weight pruning rate.
Then the entire filter/neuron whose sparsity is larger than the filter/neuron pruning rate $\gamma_F$ or $\gamma_N \in(0,1)$ is set to zero.
In this way, a large amount of entire filters/neurons are pruned, leading to a compact inference model.

\section{Algorithmic Experiments}
\label{sec:SWresult}
To evaluate the proposed approach, we present experimental results in this section. We perform experiments on several modern DNN structures (LeNet~\cite{lecun1998gradient}, VGG-Net~\cite{simonyan2014very}, ResNet~\cite{he2016deep}) and representative datasets (MNIST~\cite{lecun1998gradient}, CIFAR-10, CIFAR-100~\cite{krizhevsky2009learning}, SVHN~\cite{netzer2011reading}).
\begin{table}[!t]
\centering
\caption{Evaluation of the performance on MNIST.}
\label{table:overallresult-mnist}
\resizebox{\columnwidth}{!}{ 
\begin{tabular}{|l | r | rr | rr|}
\hline
 Method                             & Accuracy      & FLOPs                        & Pruned            & Param.          & Pruned                \\ \hline
  LeNet5-Baseline                          & 99.29             & 4.59M            & --                & 431K            & --                    \\ 
                                          Pruning~\cite{hu2016network}      & 99.26             & 0.85M            & 81.5\%            & 112K            & 74.0\%                       \\
                                          Pruning~\cite{han2015learning}    & 99.23             & 0.73M            & 84.0\%            & 36K             & 92.0\%                     \\
                                          CGaP                              & \textbf{99.36}   & \textbf{0.44M}    & \textbf{90.4\%}   & \textbf{8K}     & \textbf{98.1\%}       \\ 
\hline                                          
\end{tabular}
}

\end{table}
\begin{table}[!t]
\centering
\caption{Evaluation of the performance on CIFAR-100.  `NA' means `not available' in the original paper.}
\label{table:overallresult-CIFAR100}
\resizebox{\columnwidth}{!}{ 
\begin{tabular}{|l | r | rr | rr|}
\hline
 Method                             & Accuracy      & FLOPs                        & Pruned            & Param.          & Pruned                \\ \hline
VGG19-Baseline                           &  72.63            & 797M             & --                & 20.4M           & --                          \\
                                        Pruning~\cite{liu2018rethinking}   & 71.85             & NA                           & --                & 10.1M           & 50.5\%                            \\
                                        Pruning~\cite{liu2017learning}     & 72.85             & 501M             & 37.1\%            & 5.0M            & 75.5\%                            \\
                                        CGaP                               & \textbf{73.00}    & \textbf{373M}    &  \textbf{53.2\%}  &\textbf{4.3M}   &\textbf{ 78.9\%}        \\                                   
 \hline
\end{tabular}
}

\end{table}

\begin{table}[!t]
\centering
\caption{Evaluation of the performance on SVHN. }
\label{table:overallresult-svhn}
\resizebox{\columnwidth}{!}{ 
\begin{tabular}{|l | r | rr | rr|}
\hline
 Method                             & Accuracy      & FLOPs                        & Pruned            & Param.          & Pruned                \\ \hline
VGG19-Baseline                           & 96.02             & 797M            & --                & 20.4M           & --                     \\
                                        Pruning~\cite{liu2017learning}     & 96.13             & 398M             & 50.1\%            & 3.1M            & 84.8\%                              \\
                                        CGaP                               & \textbf{96.25}    & \textbf{206M}    & \textbf{74.2\%}   & \textbf{2.9M}   &\textbf{ 85.8\%}    \\                                
 \hline
\end{tabular}
}

\end{table}

\begin{table}[t!]
\centering
\caption{Evaluation of the performance on CIFAR-10.}
\label{table:overallresult-CIFAR10}
\resizebox{\columnwidth}{!}{ 
\begin{tabular}{|l | r | rr | rr|}
\hline
 Method                             & Accuracy      & FLOPs                        & Pruned            & Param.          & Pruned                \\ \hline
 VGG16-Baseline                          &  93.25            & 630M            & --                & 15.3M           & -- \\                      
                                        Pruning~\cite{haoli}              & 93.40             & 410M             & 34.9\%            & 5.4M            & 64.7\%                           \\
                                     CGaP                              & \textbf{93.59}    & \textbf{280M}    & \textbf{56.2\%}   & \textbf{4.5M}   &\textbf{ 70.6\%}         \\ \hline

 ResNet-56-Baseline                        &   93.03           & 268M             &   --              & 0.85M           & --                          \\ 
                                       Pruning~\cite{liu2018rethinking}   &  92.56           & 182M             & 32.1\%            & 0.73M           & 14.1\%                           \\
                                       Pruning~\cite{he2018amc}   & 90.20 & \textbf{134M} & \textbf{50.0\%}  & NA & -
                                       \\
                                       CGaP                               & \textbf{93.20}    & 181M    & 32.5\%   & \textbf{0.53M}  & \textbf{37.6\%}         \\\hline
ResNet-110-Baseline                      & 93.34             & 523M            & --                 & 1.72M          & --                           \\
                                      Pruning~\cite{haoli}               & 93.11            & 310M             & 40.7\%             & 1.16M          & 32.6\%                              \\
                                      Pruning~\cite{he2018soft}          & \textbf{93.52}    & 300M             & 40.8\%             & NA             & --                                    \\
                                      
                                      CGaP                               & 93.43             & \textbf{192M}    & \textbf{63.3\%}    & \textbf{0.62M} & \textbf{64.0\%}            \\                                     
 \hline
\end{tabular}
}

\end{table}

\subsection{Training Setup}

\paragraph{Network structures}  The LeNet-5 architecture consists of two sets of convolutional, ReLU~\cite{nair2010rectified} and max pooling layers, followed by two fully-connected layers and finally a softmax classifier.
The VGG-16 and VGG-19 structures we use have the same convolutional structure as~\cite{simonyan2014very} but are redesigned with only two fully-connected to be fairly compared with the pruning-only method~\cite{haoli}. Therefore, the VGG-16 (VGG-19) has 13 (16) convolutional layers, each is followed by a batch normalization layer~\cite{ioffe2015batch} and a ReLU activation.
The structures of ResNet-56 and ResNet-110 follow~\cite{haoli}. Each convolutional layer is followed by a batch normalization layer and ReLU activation.
During the training, the depth of the networks remains constant since CGaP does not touch the depth of the network, but the width of each layer changes.

\begin{figure*}[t!]
\begin{center}
\includegraphics[width=0.9\textwidth]{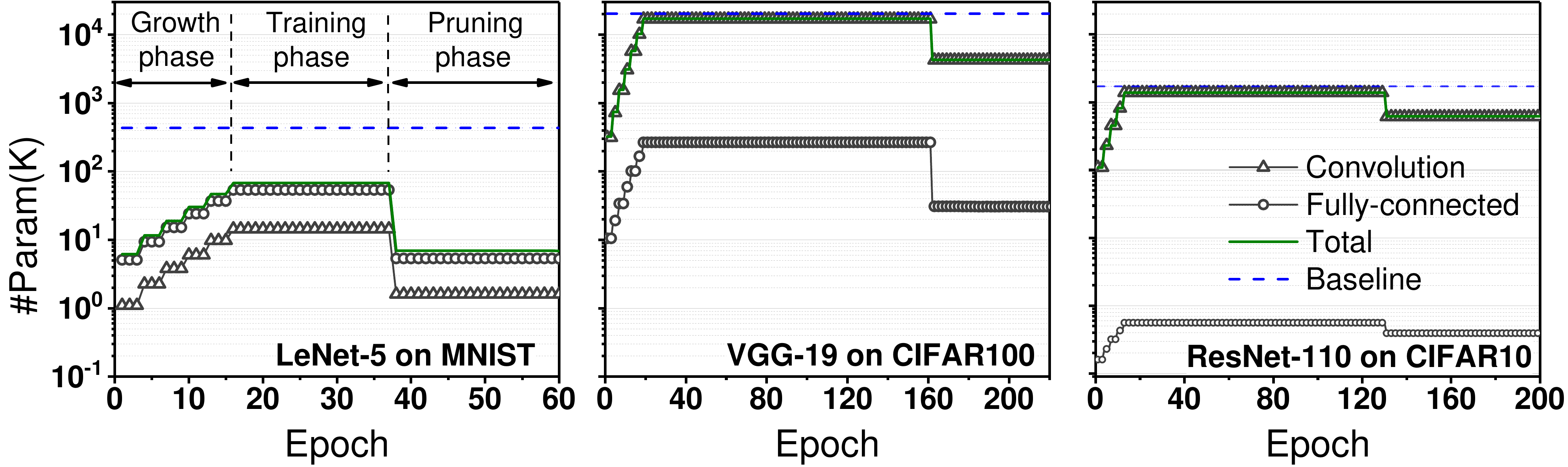}
\end{center}
\caption{Number of parameters during training, plotted at the end of each epoch. In the beginning, the model size increases gradually due to the growth. After the growth ends and several epochs of training on the peak model, one drop can be observed after the first pruning. There are several iterations of pruning at a frequency of 1.}
\label{fig:param}
\end{figure*}

\begin{figure*}[t!]
\begin{center}
\includegraphics[width=0.86\textwidth]{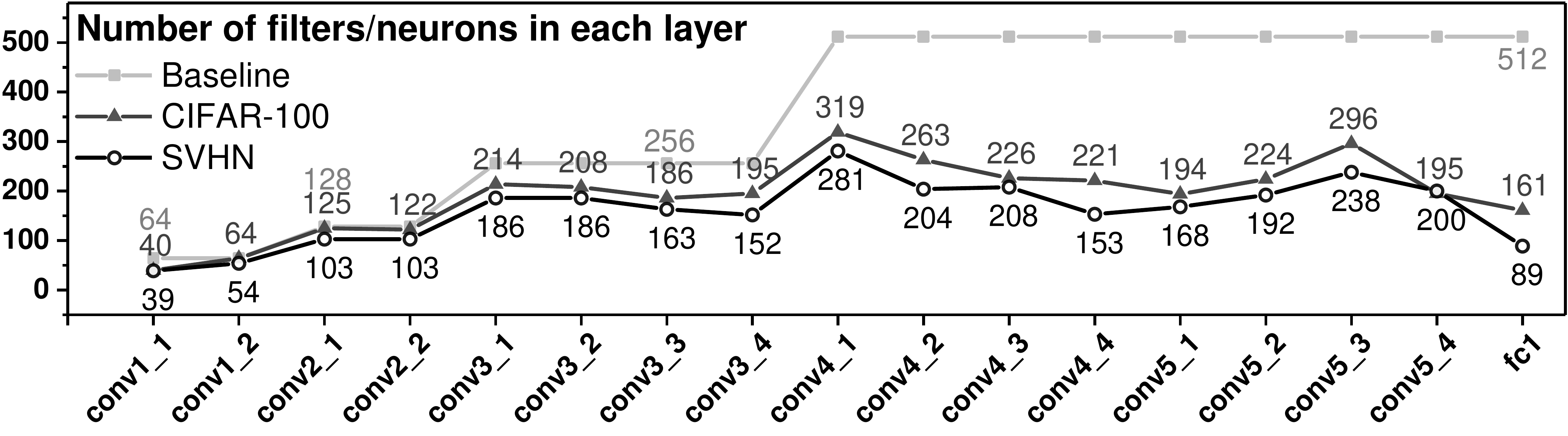}
\end{center}
\caption{The VGG-19 structures learned by CGaP on CIFAR-100 and SVHN datasets. The shared Y-axis for three sub-./ is the number of parameters of the model.}
\label{fig:layer_size}
\end{figure*}

Note that in the following text, we denote the full-size models trained from scratch without sparsity regularization as `baseline' models. The three-step pruning schemes that remove weights or filters but do not execute network growth are denoted as `pruning-only' models.

\paragraph{Datasets}
MNIST is a handwritten digit dataset in grey-scale (i.e., one color channel) with 10 classes from digit 0 to digit 9. It consists of 60,000 training images and 10,000 testing images. The CIFAR-10 dataset consists of 60,000 $32\times32$ color images in 10 classes, with 5000 training images and 1000 testing images per class. The CIFAR-100 dataset has 100 classes, including 500 training images and 100 testing images per class. The Street View House Number (SVHN) is a real-world color image dataset that is resized to a fixed resolution of $32\times32$ pixels. It contains 73,257 training images and 26,032 testing images.

\paragraph{Hyper-parameters} 
We set the learning rate to be 0.1 and divide by 10 for every 30\% of the training epochs. We train our model using Stochastic Gradient Descent (SGD) with a batch size of 128 examples, a momentum of 0.9, and a weight decay of 0.0005. The loss function is the cross-entropy loss with softmax function. We train 60, 200, 220 and 100 epochs on MNIST, CIFAR-10, CIFAR-100 and SVHN datasets, respectively. In the growth phase, we have hyper-parameters set as  follows: the growth stopping condition $\tau_{capa.}=O_{1,baseline}$, i.e., the growth stops at the $t$-th growth if the number of filters in the $(t+1)$-th growth is larger than the baseline model. The growth ratio $\beta$ is set as 0.6. The growth frequency $f_{growth}$ is set as 1/3. The scaling factor $\sigma$ in Eq.~\ref{math:initial1} to Eq.~\ref{math:initial4-fc} is set to 0.5 and $\mu$ is 0.1. The pruning frequency $f_{pruning}$ is set to be 1. The setting of the weight pruning rate $\gamma_W$ follows~\cite{han2015learning}, \cite{haoli} and \cite{liu2017learning} for LeNet-5, VGG-Net and ResNet, respectively. $\gamma_F$ and $\gamma_N$ is set to be same as $\gamma_W$.

\paragraph{Framework and platform}
The experiments are performed with PyTorch~\cite{paszke2017automatic} framework on one NVIDIA GeForce GTX 1080 Ti platform. It is worth mentioning that experiments performed with different frameworks may have variation in accuracy and performance.  Thus, to have a fair comparison among CGaP, baseline and pruning-only methods, all the results in Table ~\ref{table:overallresult-mnist},~\ref{table:overallresult-CIFAR100}, ~\ref{table:overallresult-svhn} and~\ref{table:overallresult-CIFAR10} are obtained from experiments with PyTorch framework.

\subsection{Performance Evaluation}
With training setup as aforementioned, we perform experiments on several datasets with modern DNN architectures. 
In Table~\ref{table:overallresult-mnist}, Table~\ref{table:overallresult-CIFAR100}, Table~\ref{table:overallresult-svhn} and Table~\ref{table:overallresult-CIFAR10}, we summarize the performance attained by CGaP on MNIST, CIFAR-100, SVHN, and CIFAR-10 datasets, respectively. 
To be specific, the second column `Accuracy' denotes the inference accuracy in percentage achieved by the baseline model, the up-to-date pruning-only approaches and CGaP approach, respectively.

The column `FLOPs' represent the calculated number of FLOPs of a single inference pass. The calculation of FLOPs follows the method described in~\cite{molchanov2016pruning}. Fewer FLOPs means lower computation cost in one inference pass. The neighboring column, `Pruned', represents the reduction of FLOPs in the compressed model as compared to the baseline model. 
The column `Param.' stands for the number of parameters of the inference model. Fewer parameters promise a smaller model size.  The last column, `Pruned', denotes the percentage pruned in parameters compared to the baseline.
Larger pruned percentage implies fewer computation operations and more compact model.
The best result of each column is highlighted in bold.

The results shown in Table~\ref{table:overallresult-mnist} to~\ref{table:overallresult-CIFAR10} prove that CGaP outperforms the previous pruning-only approaches in accuracy and model size. For instance, as displayed in Table~\ref{table:overallresult-CIFAR10}, on ResNet-56, our CGaP approach achieves 93.20\% accuracy with 32.5\% reduction in FLOPs and 37.6\% reduction in parameters, while the up-to-date pruning-only method~\cite{liu2018rethinking} that deals with static structure only reaches 92.56\% accuracy with 32.1\% reduction in FLOPs and 14.1\% reduction in parameters. On ResNet-110, though~\cite{he2018soft} achieves 0.09\% higher accuracy than CGaP, CGaP overwhelms it by trimming 22.5\% more FLOPs.

\subsection{Visualization of the dynamic structures}

Fig.~\ref{fig:param} presents the dynamic model size during CGaP training. During the growth phase, the model size continuously increases and reaches a peak capacity. When the pruning phase starts, the model size drops.

Furthermore, the sparsity achieved by CGaP is structured. In other words, large amounts of filters and neurons are entirely pruned. For instance, the baseline LeNet-5 without sparsity regularization has 20, 50 filters in conv-layer 1 and conv-layer 2, 500 and 10 neurons in fc-layer 1 and fc-layer 2, denoted as [20-50-500-10] (number of filters/neurons in [conv1-conv2-fc1-fc2]). The model achieved by CGaP contains only 8, 17 filters and 23, 10 neurons, denoted as [8-17-23-10]. Compared to baseline results, CGaP significantly decreases 60\%, 66\%, 95.4\% units for each layer (the output layer should remain the same as the number of classes all the time). In this case, the pruned filters and neurons are skipped in the inference pass and thus accelerating the computation pipeline on hardware.

Another example is provided in Fig.~\ref{fig:layer_size}, which visualizes the VGG-19 structures from CGaP as well as the baseline structure on two different tasks. In the baseline model, the width (number of filters/neurons) of each layer is abundant, from 64 filters (the bottom conv-layers) to 512 filters (the top conv-layers). The baseline VGG-19 structure is designed to have a large enough size in order to guarantee the learning capacity. However, it turns out to be redundant, as proved by the structure that CGaP generated: $37.7\%$ to $82.6\%$ filters are pruned out in each layer. Meanwhile, in the baseline model, the top conv-layers are designed to have more filters than the bottom layers, but CGaP shows that it is not always necessary for top layers to have a relatively large size.

\begin{figure}[t!]
\begin{center}
\includegraphics[width=0.8\columnwidth]{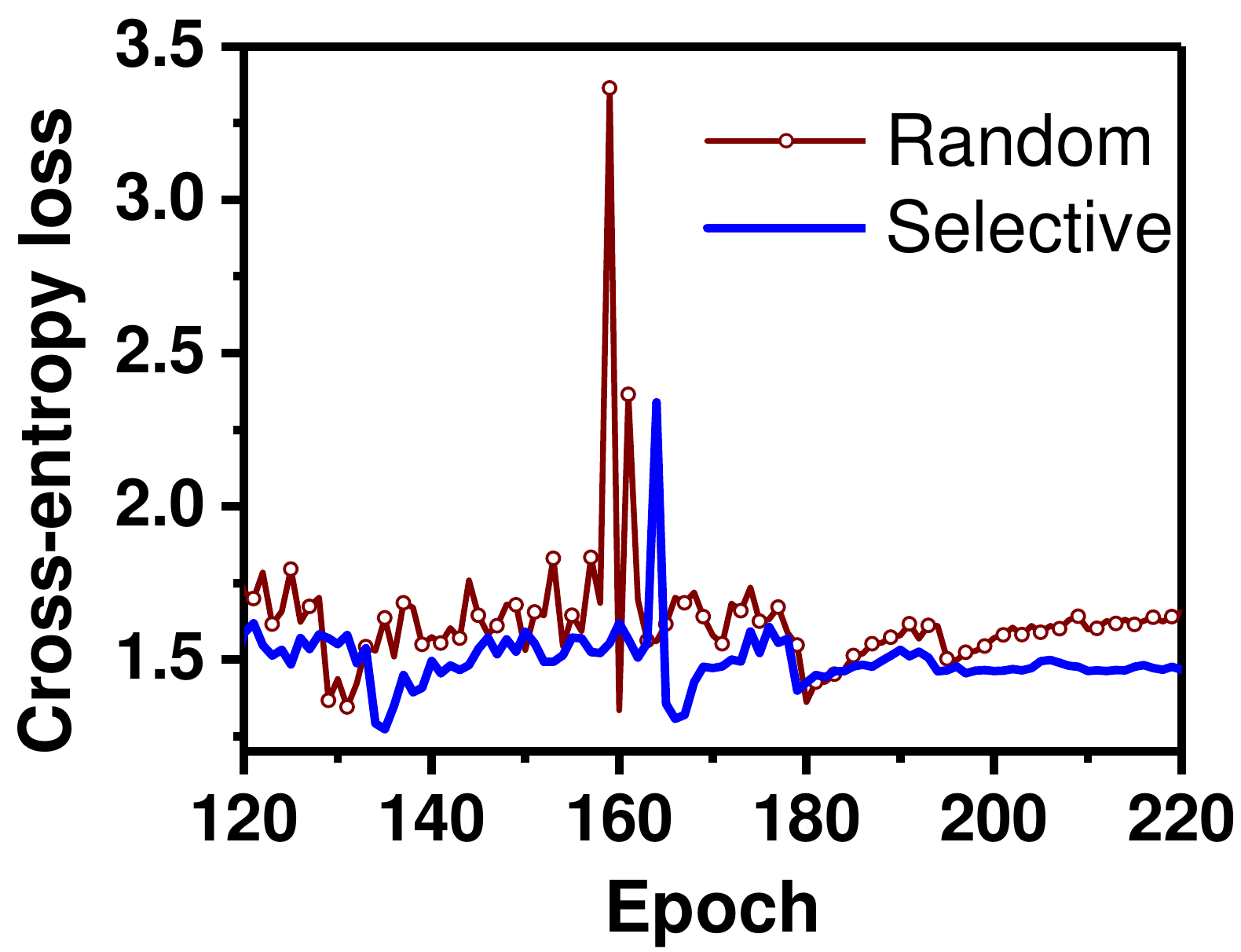}
\end{center}
\caption{Saliency-based growth outperforms random growth. The loss is monotonically decreasing from epoch 0 to 220 with small glitches. Here we zoomed in from epoch 120 to 220 to show the loss at the end of the training.}
\label{fig:rdm_select}
\end{figure}

\subsection{Validating the saliency-based growth} \label{sec:IV-D}
Fig.~\ref{fig:rdm_select} validates the efficacy of our saliency-based growth policy.
Selective growth, which emphasizes the important units according to the saliency score, has lower cross-entropy loss than randomly growing some units. The spiking in Fig.~\ref{fig:rdm_select} is caused by the first iteration of pruning and this loss is recovered by the following iterative fine-tuning. In selective growth, this loss is  $1.4\times$ lower than that in random growth. This phenomenon supports our argument that selective growth assists the pruning phase. The detailed understanding of growth will be further discussed in Section~\ref{section:discussion}.

To summarize the results from the algorithm simulations, the proposed CGaP approach:
\begin{itemize}
    \item Largely compresses the model size by $37.6\%$ (ResNet-56) to $98.1\%$ (LeNet-5) for representative DNN structures.
    \item Decreases the inference cost, to be specific, number of FLOPs, by $32.5\%$ (ResNet-56) to $90.4\%$ (LeNet-5) on various datasets.
    \item Does not sacrifice accuracy and even improves accuracy.
    \item Outperforms the state-of-the-art pruning-only methods that deal with fixed structures.
\end{itemize}

\begin{figure}[t!]
\centering
\subfigure[Comparison of three schemes in buffer access energy (pJ) for VGG-16 on CIFAR-10, VGG-19 on CIFAR-100, ResNet-56 and ResNet-110 on CIFAR-10.]{
\includegraphics[width=0.95\columnwidth]{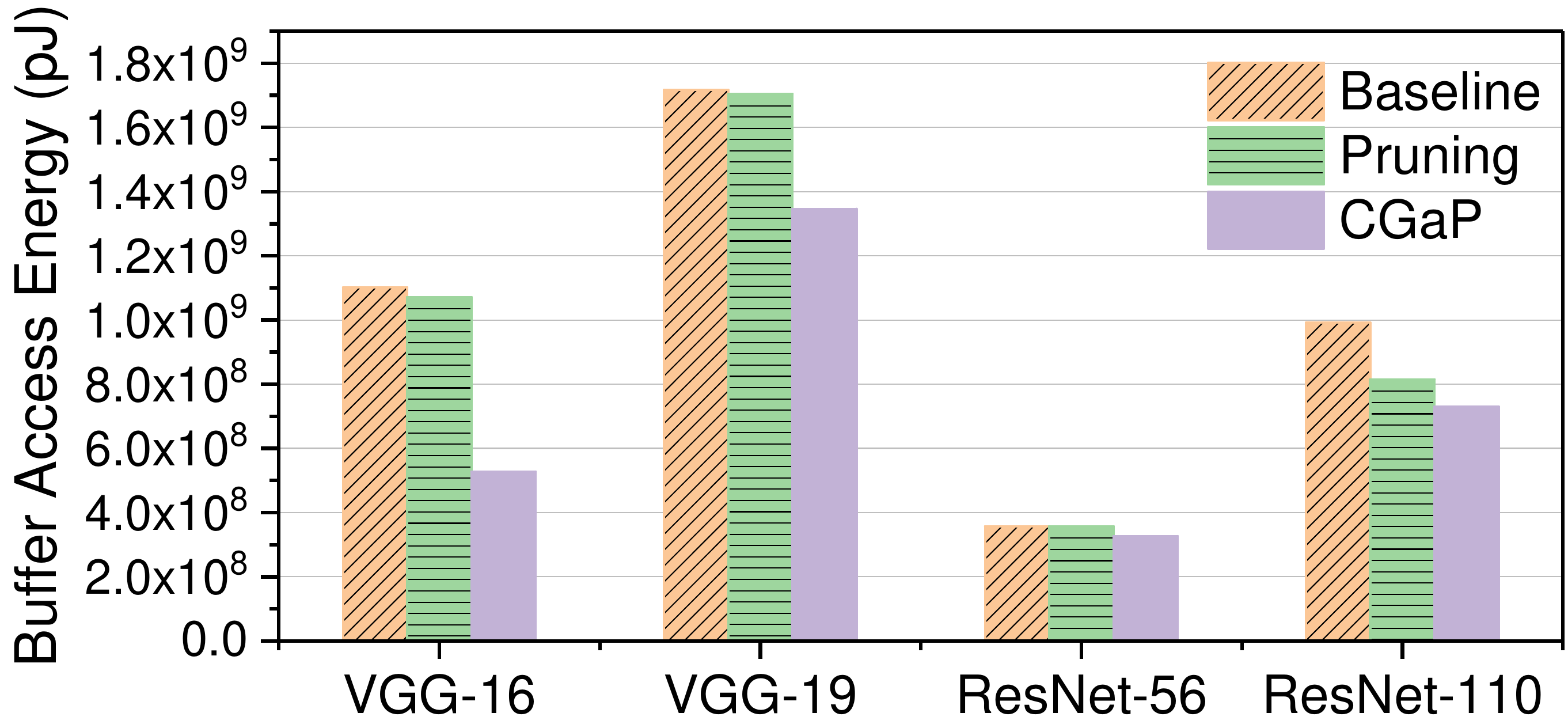}
\label{fig:buffer}
}
\quad
\subfigure[Comparison of three schemes in DRAM access energy (pJ) for VGG-16 on CIFAR-10, VGG-19 on CIFAR-100, ResNet-56 and ResNet-110 on CIFAR-10.]{  
\includegraphics[width=0.95\columnwidth]{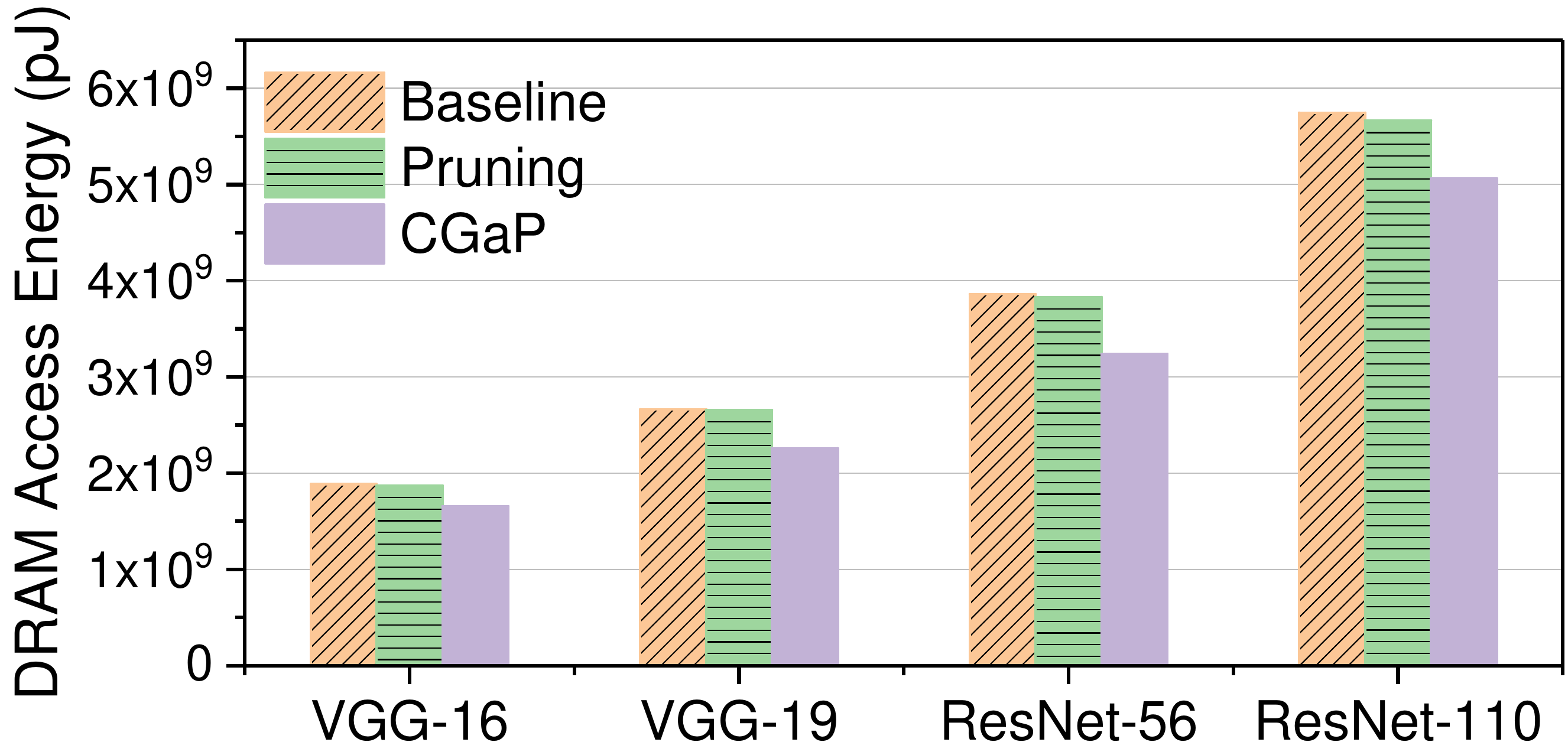}
\label{fig:dram}
}
\quad
\subfigure[Comparison of three schemes in on-chip inference latency (ms) for VGG-16 on CIFAR-10, VGG-19 on CIFAR-100,  ResNet-56 and ResNet-110 on CIFAR-10.]{  
\includegraphics[width=0.95\columnwidth]{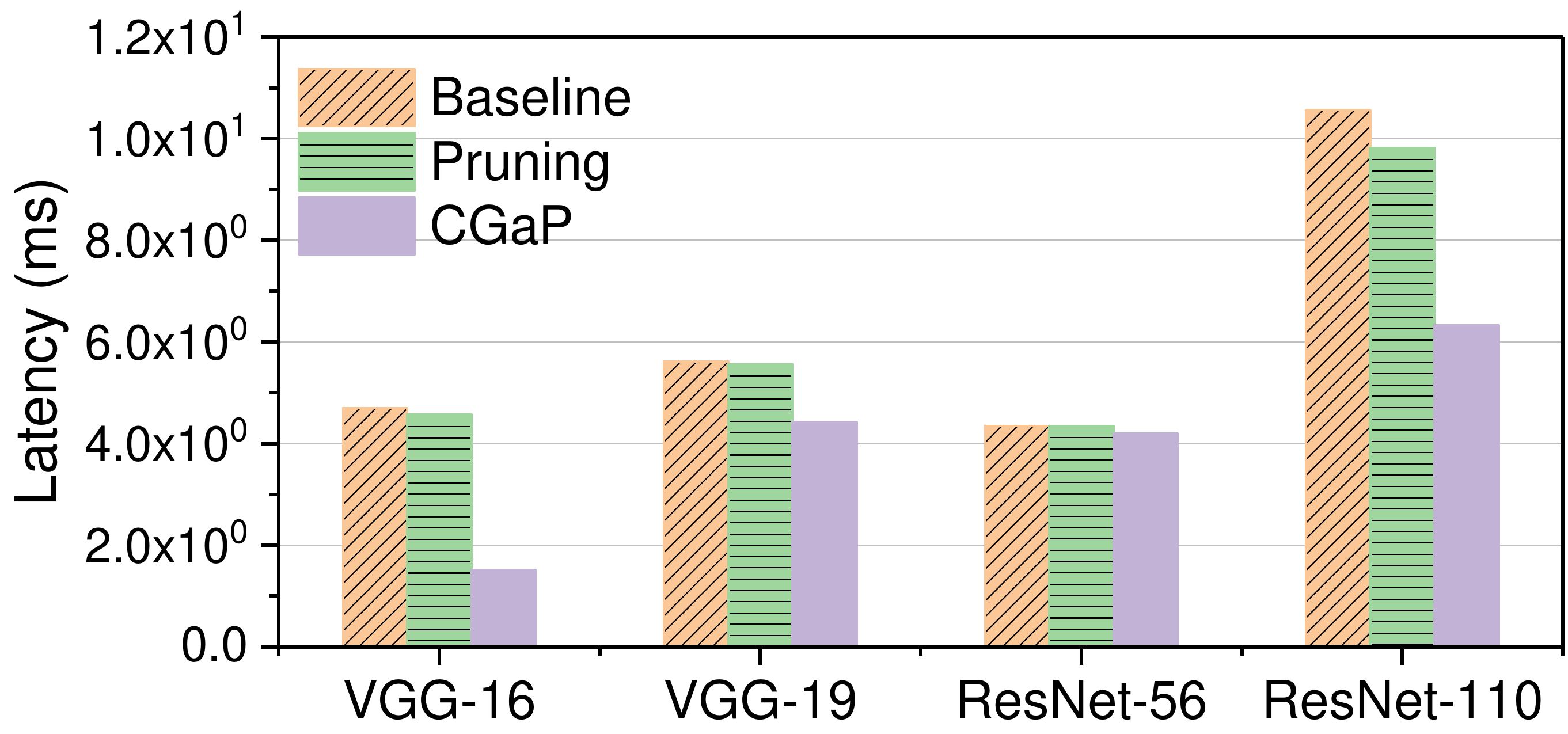}
\label{fig:latency}
}
\caption{Estimation on FPGA performance model.}
\end{figure}

\section{Experiments on FPGA simulator}
\label{sec:HWresult}
The results above demonstrate that CGaP generates an accurate and small inference model.  In this section, we further evaluate the on-chip inference cost of the generated models and compare CGaP with previous non-structured pruning~\cite{han2015learning}. As CGaP achieves structured sparsity, CGaP outperforms the previous work on non-structured pruning in hardware acceleration and power efficiency. We validate this by performing the estimation of buffer access energy, DRAM access energy and latency using the performance model for FPGA~\cite{ma2019performance}. 

\subsection{Overview of the FPGA simulator}
\cite{ma2019performance} is a high-level performance model designed to estimate the number of external and on-chip memory access, as well as the latency. The resource costs are formulated by the acceleration strategy as well as the design variables that control the loop tiling and unrolling. The performance model has been validated across several modern DNN algorithms in comparison to on-board testings on two FPGAs, with the differences within $3\%$~\cite{ma2019performance}. 

In the following experiments, the setup follows: the pixels and weights are both 16-bit fixed point, the data width of DRAM controller is $512$ bits, the accelerator operating frequency is $300$ MHz, and the DRAM bandwidth is $19.2$ GB/second. The parameters related to loop tiling and unrolling follow the setting in~\cite{ma2019performance}.

\subsection{Results from FPGA performance model}
The on-chip and external memory access energy across VGG-16, VGG-19, ResNet-56 and ResNet-110 is displayed in Fig.~\ref{fig:buffer} and Fig.~\ref{fig:dram}, respectively. The inference latency is shown in Fig.~\ref{fig:latency}.  Though the models generated from weight magnitude pruning and CGaP have the same sparsity, CGaP outperforms non-structured magnitude weight pruning in hardware efficiency and acceleration. 
For example, with the same setup of the pruning ratio during training, magnitude weight pruning decreases $1.0\%$ on-chip access energy, $1.0\%$ DRAM access energy and $0.8\%$ latency for VGG-19 on CIFAR-100, while the CGaP achieves $21.6\%$, $15\%$, and $21.1\%$ reduction.
The non-structured weight pruning~\cite{han2015learning} is able to improve the power and latency efficiency in comparison to baseline. However, the improvement is limited. In contrast, CGaP achieves significant acceleration and energy reduction. 
The reason is that the non-structured sparsity, i.e., scattered weight distribution, leads to irregular memory access that weakens the acceleration on hardware in a real scenario.

\section{Discussion}
\label{section:discussion}
In Section~\ref{sec:SWresult} and~\ref{sec:HWresult}, the performance of CGaP has been comprehensively evaluated on algorithm platforms and hardware platforms. In this section, we  provide a more in-depth understanding of the growth to explain why selective growth is able to improve the performance from the traditional pipelines. Furthermore, we provide a thorough ablation study to validate the robustness of the proposed CGaP method.

\begin{figure}[t!]
\begin{center}
\includegraphics[width=\columnwidth]{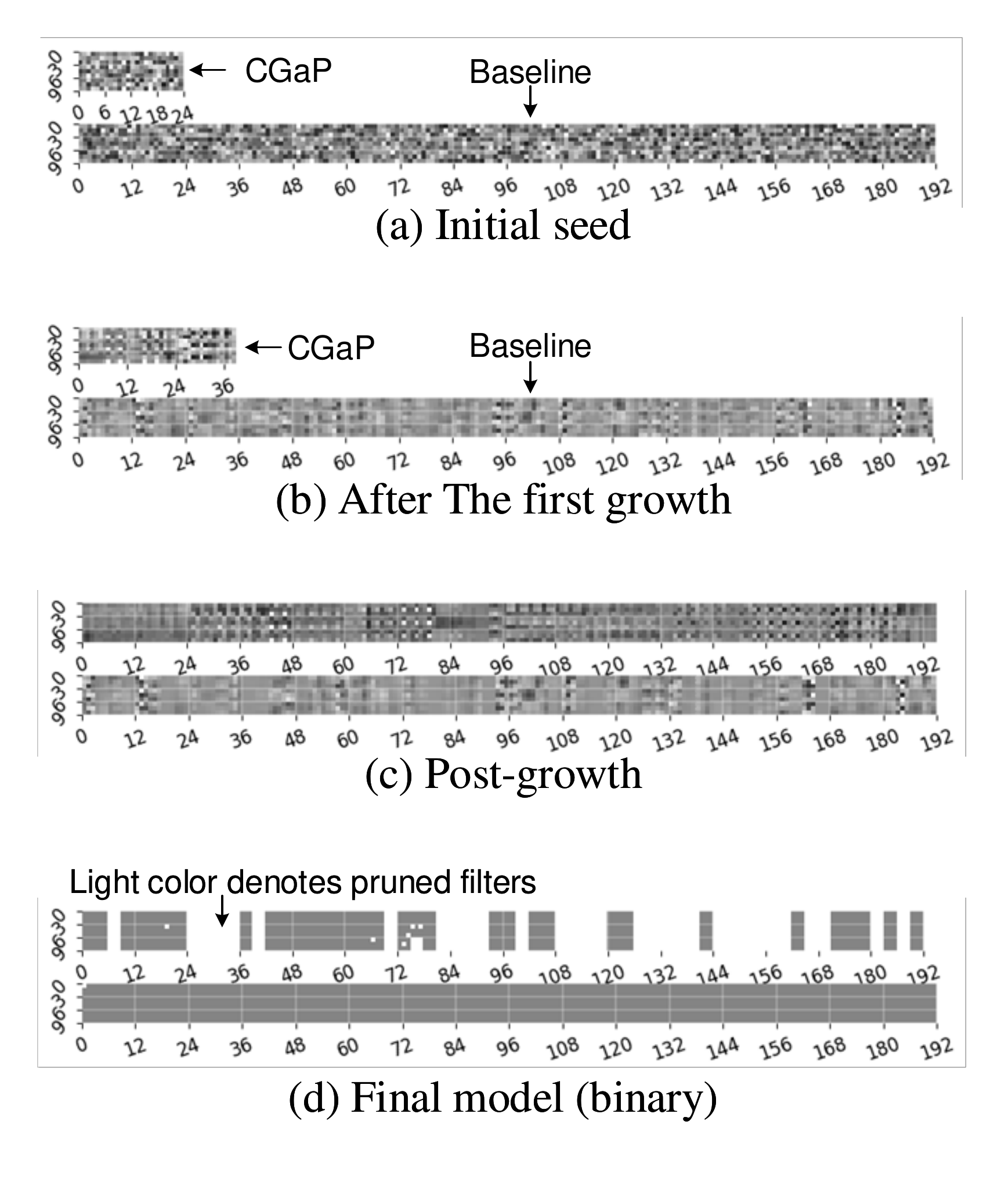}
\end{center}
\caption{Visualization of the filters in conv1\_1 in VGG-19 on CIFAR-100 at four specific moments (a-d). Inside each figure, the top bar is CGaP model and the bottom bar is baseline model.  X-axis is the index of output-wise weights and Y-axis is the index of input-wise weights.}
\label{fig:heatmap}
\end{figure}

\begin{table*}[t!]
\centering
\resizebox{0.8\textwidth}{!}{ 
\begin{tabular}{|c|c|r|r|r|r|r|r|}\hline
\multicolumn{2}{|c|}{\textbf{Initial seeds}} & \textbf{`2'} & \textbf{`4'} & \textbf{`6'} & \textbf{`8'} & \textbf{`10'} & \textbf{`12'} \\ \hline 
\multirow{5}{*}{\#filters} & conv1\_$n$ & 2 & 4 & 6 & 8 & 10 & 12\\
 & conv2\_$n$ & 4 & 8 & 12 & 16 & 20 & 24\\
 & conv3\_$n$ & 8 & 16 & 24 & 32 & 40 & 48\\
 & conv4\_$n$ & 16 & 32 & 48 & 64 & 80 & 96\\ 
 & conv5\_$n$ & 16 & 32 & 48 & 64 & 80 & 96\\ \hline
\#param & Initial (M) & 0.01 & 0.06 & 0.13 & 0.23 & 0.36 & 0.53 \\ \hline\hline
\multicolumn{2}{|c|}{Testing accuracy*} & -0.69\% & -0.2\% & -0.16\% & +0.37\% & +0.04\% & 0.29\% \\\hline
\multicolumn{8}{|l|}{*Relative accuracy of the final VGG-19 model on CIFAR-100 as compared  to the baseline.} \\ \hline

\end{tabular}
}
\caption{The impact of various structures and and sizes of the initial seed of VGG-19.}
\label{table: seed}
\end{table*}

\paragraph{Understanding the growth}  

Fig.~\ref{fig:heatmap} illustrates a visualization of the weights in the bottom conv-layer (conv1\_1) in VGG-19, at the moment of initialization, after the first growth, after the last growth and when training ends. Inside each figure, the upper bar is the CGaP model, whose size varies at different training moments. The lower bar is from the baseline model, whose size is static during training. At the initialization moment (Fig.~\ref{fig:heatmap}(a)), CGaP model only has 8 filters in this layer while the baseline model has 64 filters. Then the number of filters grows to 13 after one iteration growth (Fig.~\ref{fig:heatmap}(b)), meaning the most important 5 filters are selected and added. It is clear that the pattern in Fig.~\ref{fig:heatmap}(b) is more active than that in (a), indicating the filters have already fetched effective features from the input images. More important, along with the growing, the pattern in CGaP model becomes more structured than that in the baseline model, as shown in Fig.~\ref{fig:heatmap}(c). Benefiting from this well-structured pattern, our CGaP model has higher learning accuracy than the baseline model. From Fig.~\ref{fig:heatmap}(c) to Fig.~\ref{fig:heatmap}(d), relatively unimportant filters are removed, and important ones are kept. We observe that most of the filters that are favored by the growth, such as filters at index 36, 48, 72, 96 in Fig.~\ref{fig:heatmap}(c), are still labeled as important filters in Fig.~\ref{fig:heatmap}(d) even after a long training process between the growth phase and the pruning phase. Leveraging the growth policy, the model is able to recover quickly from the loss caused by pruning (the spiking in Fig.~\ref{fig:rdm_select}).

\begin{figure}[t!]
\begin{center}
\includegraphics[width=0.8\columnwidth]{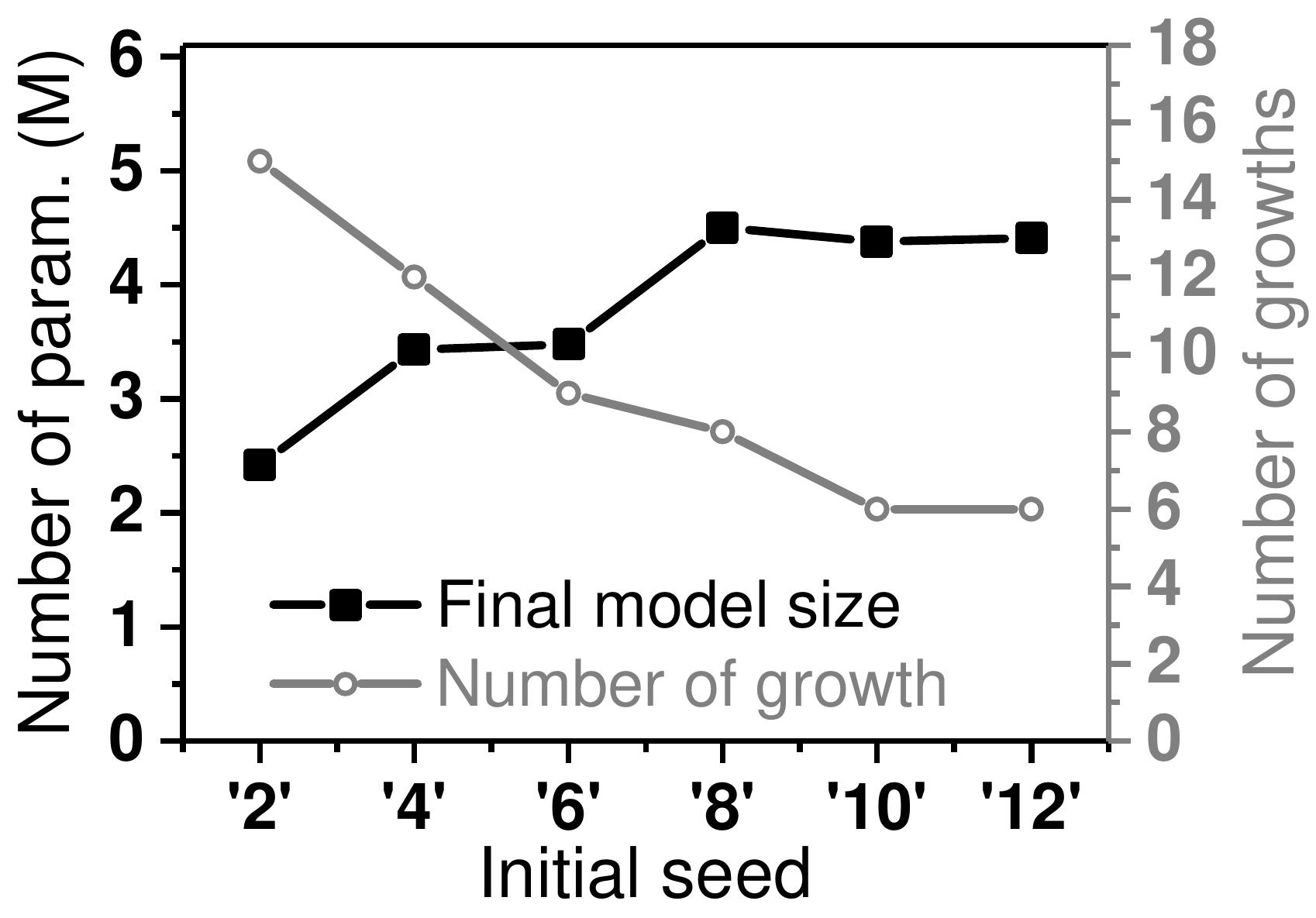}
\end{center}
\caption{A larger seed leads to a larger final model but fewer iterations in the growth phase.}
\label{fig:seed}
\end{figure}

\paragraph{Robustness of the seed} 
The performance of CGaP is stable under the variation of the initial seeds. To prove this, we scan several seeds in different size and present the variation in accuracy and inference model size. The structure of 6 scanned seeds is listed in Table~\ref{table: seed}. Each seed has a different number of filters in each layer, e.g., seed `2' has 2 filters in block conv1. The size of the seeds varies from 0.01M to 0.53M.  Fig.~\ref{fig:seed} presents the final model size and the number of growth of each seed. A larger seed leads to a larger final model but requires fewer iterations of growth to reach the intended model size.  Generally speaking, there is a trade-off between the inference accuracy and the model size. Though the seed varies a lot from each other, the final accuracy is quite robust, as listed in the `Accuracy' row in Table~\ref{table: seed}. It is worth mentioning that, even though the seed `2' degrades the accuracy of $0.69\%$ from baseline, the inference model size is only 2.4M, significantly smaller than the baseline size (20.4M).

\paragraph{Robustness of the hyper-parameters}

CGaP is conditioned on a set of hyper-parameters to achieve optimal performance, while it is stable under the variation of these hyper-parameters. Empirically, we leverage the following experience to perform parameter optimization: a smaller growth rate $\beta$ for a larger seed and vice versa; threshold $\tau_{capa}$ is set based on the user's intended model size; a smaller $f_{growth}$ for a complicated dataset and vice versa; a relatively greedy growth (larger $\beta$ and $f_{growth}$) prefers a larger noise $\mu$ but smaller $\sigma$ to push the model away from sticking at a local minimum. Tuning of the pruning ratio of each layer is in a similar manner to that of the other pruning works~\cite{han2015learning}~\cite{haoli}.

In particular, we scan 121 combinations of the scaling factor $\sigma$ and noise $\mu$ in the range [0.0, 1.0] with the step=0.1 and provide the following discussion. For VGG16 on CIFAR-10, the accuracy of several corner cases are $90\%$ ($\mu$=1, $\sigma$=0, which is a case of random initialization), $89\%$ ($\mu$=1, $\sigma$=1), $84\%$ ($\mu$=0, $\sigma$=1, which is another case of mimicking its neighbor without scaling) and $10\%$ ($\mu$=0, $\sigma$=0, the training is invalid in this case), $93\%$ ($\mu$=0, $\sigma$=0, which is another case of mimicking its neighbor with scaling), $88\%$ ($\mu$=0.5, $\sigma$=0, which is another case of random initialization). The best accuracy of $93.6\%$ is under $\mu$=0.1, $\sigma$=0.5. The combinations in the zone that $\mu \in [0, 0.5]$ and $\sigma \in (0, 0.5]$ always provide \textgreater$92\%$ accuracy. To summarize, $\sigma$ impacts more than $\mu$ as $\mu$ is relatively small; $\sigma$ should not be too large and 0.5 is safe for future tasks and networks; adding a noise improves the accuracy (like from $93\%$ to $93.6\%$)  as it prevents local minimum; inheriting from the neighbor is more efficient than randomly initializing since the network is able to resume the learning right after the growth.

\section{Conclusion and Future Work}
\label{sec:conclusion}
Modern DNNs typically start training from a fixed and over-parameterized network, which leads to redundancy and is lack of structural plasticity. We propose a novel dynamic training algorithm, Continuous Growth and Pruning, that initializes training from a small network, expands the network width continuously to learn important learning units and structures and finally prunes secondary ones. The effectiveness of CGaP depends on where to start and stop the growth,  which learning unit (filter and neuron) should be added, and how to initialize the newborn units to ensure model convergence. Our experiments on benchmark datasets and architectures demonstrate the advantage of CGaP on learning efficiency (accurate and compact). We further validate the energy and latency efficiency of the inference model generated by CGaP on FPGA performance simulator. Our approach and analysis will help shed light on the development of adaptive neural networks for dynamic tasks such as continual and lifelong learning.

\section*{Acknowledgment}
This work was supported in part by C-BRIC, one of six centers in JUMP, a Semiconductor Research Corporation (SRC) program sponsored by DARPA. It was also partially supported by National Science Foundation (NSF) under CCF \#1715443.
\ifCLASSOPTIONcaptionsoff
  \newpage
\fi


\vfill
\begin{IEEEbiography}[{\includegraphics[clip,width=1in,height=1.25in,keepaspectratio]{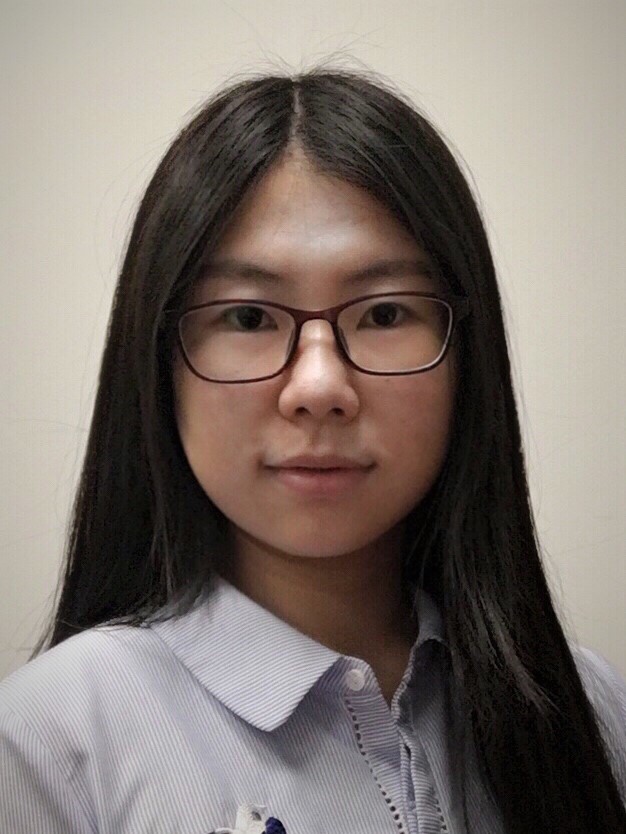}}]
{Xiaocong Du} (S'19) received her B.S. degree in control engineering from Shandong University, Jinan, China, in 2014, and the M.S. degree in electrical and computer engineering from University of Pittsburgh, Pittsburgh, PA, US, in 2016. Now, she is pursuing her Ph.D. degree in electrical engineering at Arizona State University, Tempe, AZ, USA. Her research interests include efficient algorithm and hardware co-design for deep learning, neural architecture search, continual learning, and neuromorphic computing. 
\end{IEEEbiography}
\vfill
\vfill
\begin{IEEEbiography}[{\includegraphics[clip,width=1.0in,height=1.25in,keepaspectratio]{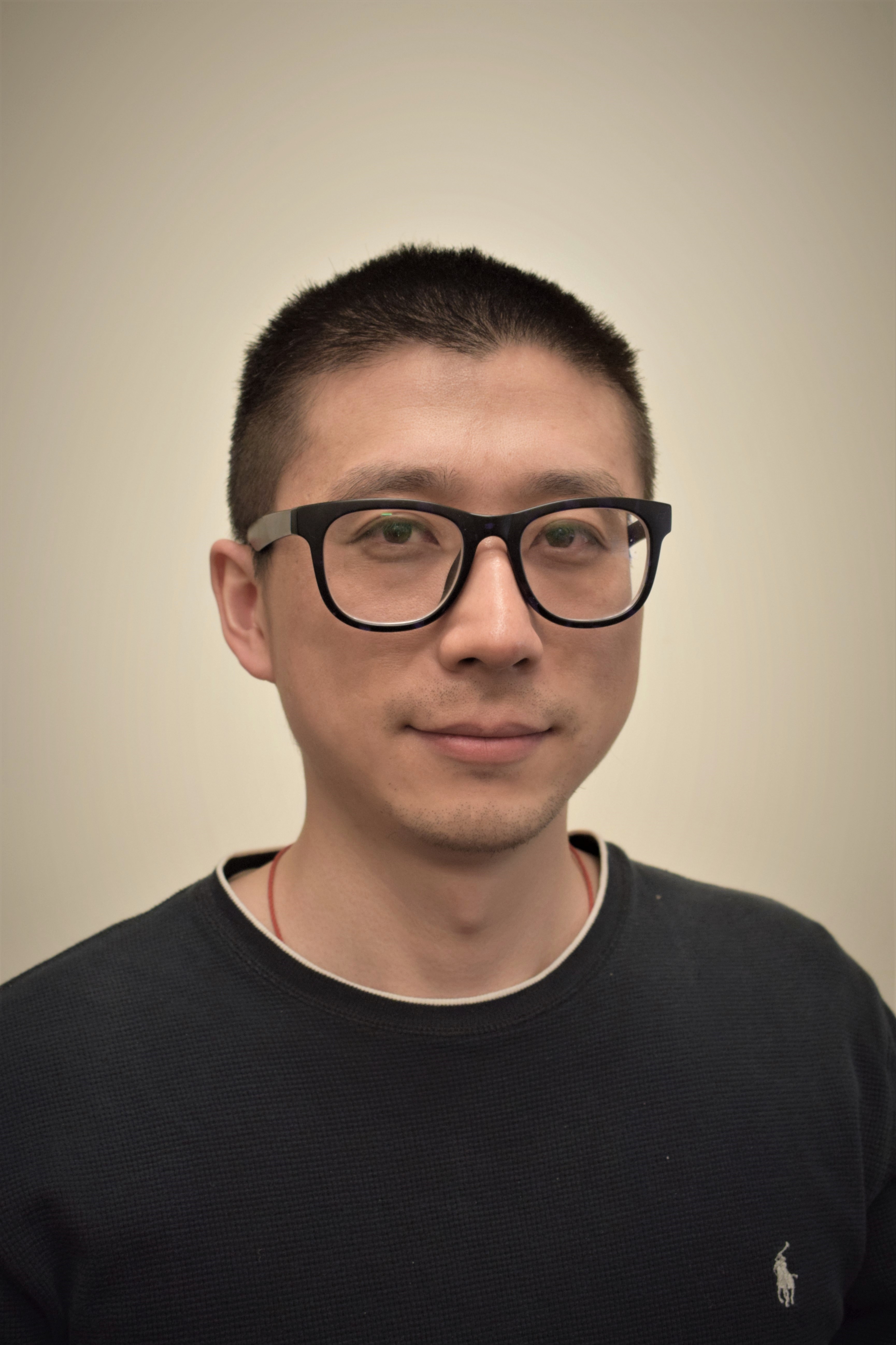}}]
{Zheng Li} (S'19) obtained his B.S. degree in electronics and information engineering from Beihang University, Beijing, China, in 2014, and the M.S. degree in electrical and computer engineering from University of Pittsburgh, Pittsburgh, PA, USA, in 2017. He is currently working towards the Ph.D. degree in computer engineering at Arizona State University, Tempe, AZ, USA.
He worked as a summer intern in Machine Learning at MobaiTech, Inc, Tempe, AZ, USA in 2018. His current research interests include algorithm design and optimization for computer vision tasks, such as object detection and autonomous driving.
\end{IEEEbiography}
\vfill
\vfill
\begin{IEEEbiography}[{\includegraphics[clip,width=1in,height=1.25in,keepaspectratio]{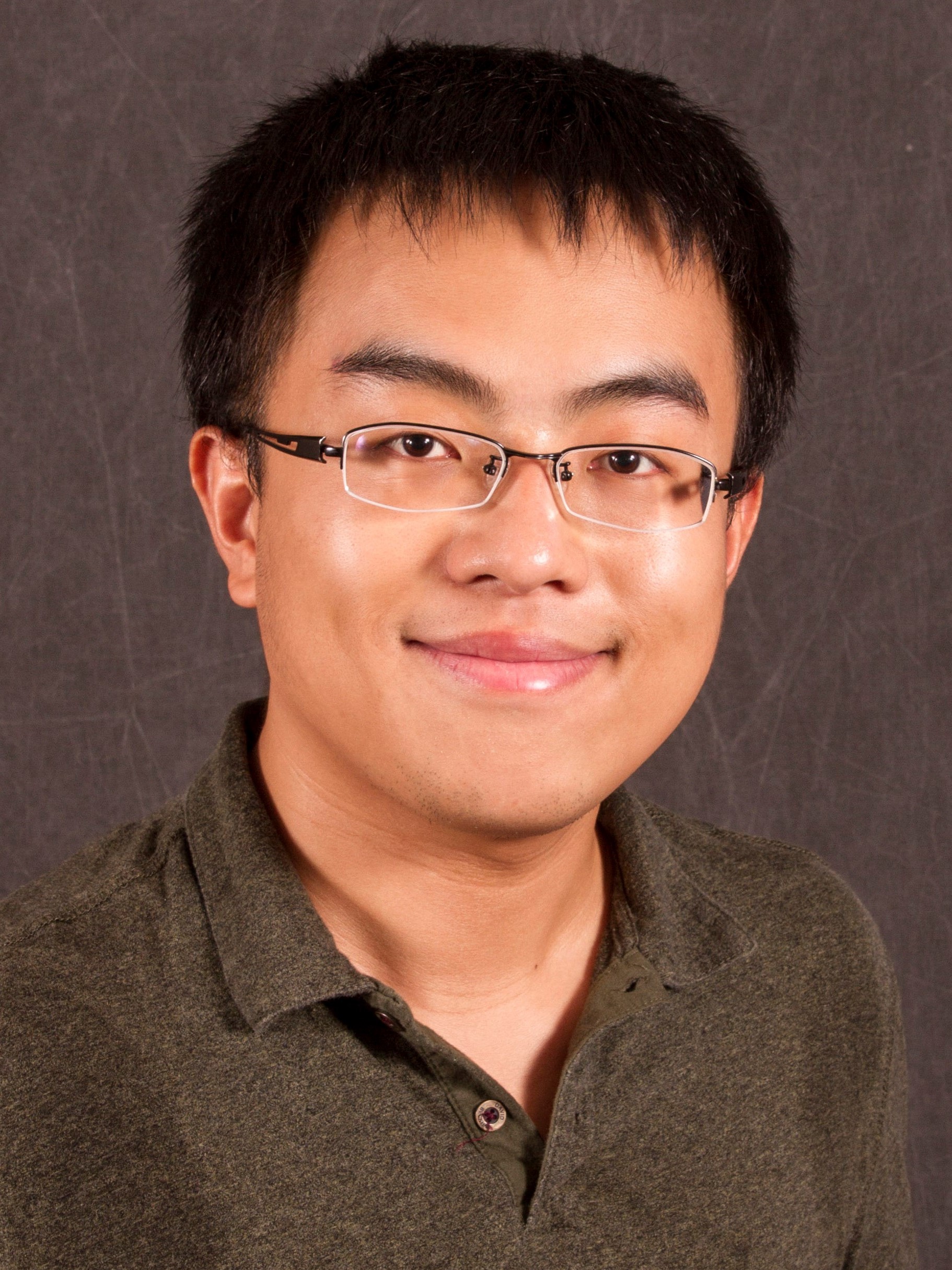}}]
{Yufei Ma} (S'16-M'19) received the B.S. degree in information engineering from Nanjing University of Aeronautics and Astronautics, Nanjing, China, in 2011, 
the M.S.E. degree in electrical engineering from University of Pennsylvania, Philadelphia, PA, USA, in 2013, and the Ph.D. degree with Arizona State University, Tempe, AZ, USA, in 2018.
His current research interests include the high-performance hardware acceleration of deep learning algorithms on digital application-specified integrated circuit and field-programmable gate array.
\end{IEEEbiography}
\vfill

\vfill
\begin{IEEEbiography}[{\includegraphics[clip,width=1in,height=1.25in,keepaspectratio]{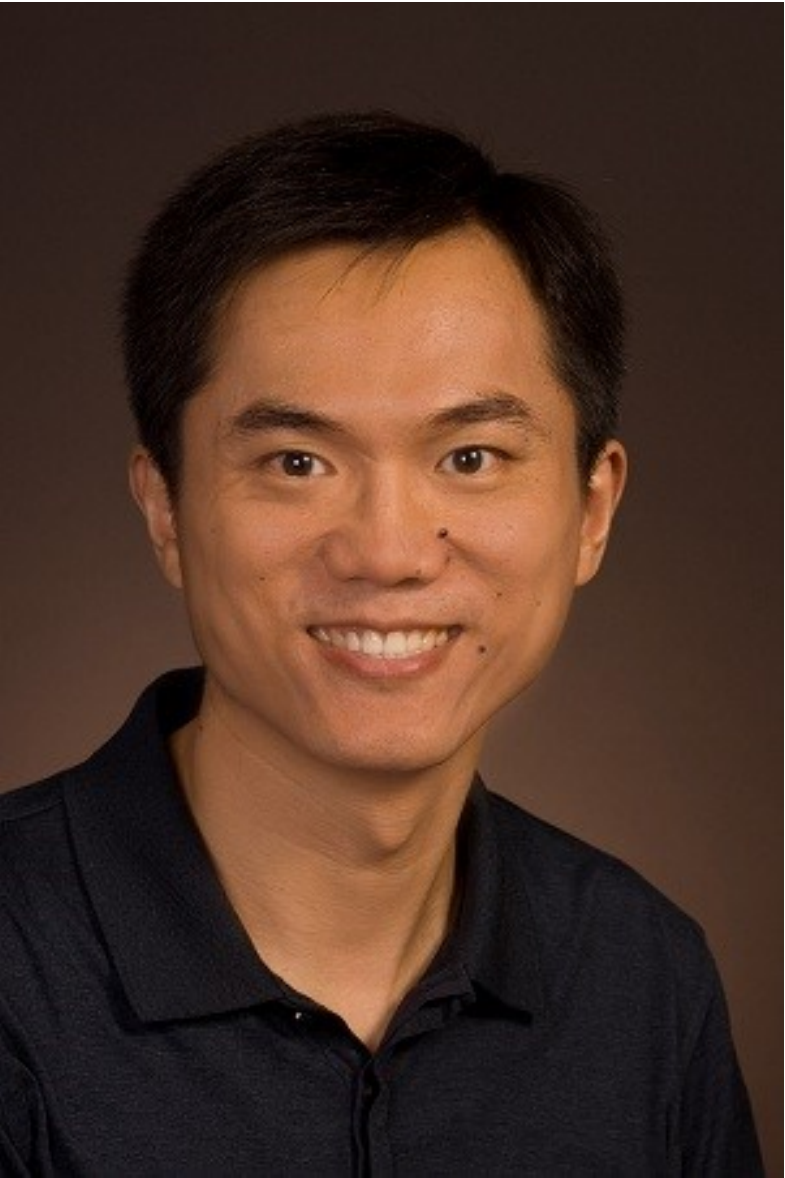}}] {Yu Cao} (S'99-M'02-SM'09-F'17) received the B.S. degree in physics from Peking University in 1996. He received the M.A. degree in biophysics and the Ph.D. degree in electrical engineering from University of California, Berkeley, in 1999 and 2002, respectively. 
He worked as a summer intern at Hewlett-Packard Labs, Palo Alto, CA in 2000, and at IBM Microelectronics Division, East Fishkill, NY, in 2001. After working as a post-doctoral researcher at the Berkeley Wireless Research Center (BWRC), he is now a Professor of Electrical Engineering at Arizona State University, Tempe, Arizona. He has published numerous articles and two books on nano-CMOS modeling and physical design. His research interests include physical modeling of nanoscale technologies, design solutions for variability and reliability, reliable integration of post-silicon technologies, and hardware design for on-chip learning.   
Dr. Cao was a recipient of the 2012 Best Paper Award at IEEE Computer Society Annual Symposium on VLSI, the 2010, 2012, 2013, 2015 and 2016 Top 5\% Teaching Award, Schools of Engineering, Arizona State University, 2009 ACM SIGDA Outstanding New Faculty Award, 2009 Promotion and Tenure Faculty Exemplar, Arizona State University, 2009 Distinguished Lecturer of IEEE Circuits and Systems Society, 2008 Chunhui Award for outstanding oversea Chinese scholars, the 2007 Best Paper Award at International Symposium on Low Power Electronics and Design, the 2006 NSF CAREER Award, the 2006 and 2007 IBM Faculty Award, the 2004 Best Paper Award at International Symposium on Quality Electronic Design, and the 2000 Beatrice Winner Award at International Solid-State Circuits Conference. He has served as Associate Editor of the IEEE Transactions on CAD, and on the technical program committee of many conferences.

\end{IEEEbiography}
\vfill



\end{document}